\documentclass{article} 
\usepackage{iclr2027_conference,times}

\usepackage{amsmath,amsfonts,bm}

\def\eqref#1{equation~\ref{#1}}
\def\Eqref#1{Equation~\ref{#1}}

\def\1{\bm{1}}

\DeclareMathAlphabet{\mathsfit}{\encodingdefault}{\sfdefault}{m}{sl}
\SetMathAlphabet{\mathsfit}{bold}{\encodingdefault}{\sfdefault}{bx}{n}

\usepackage{url}
\usepackage{graphicx}
\usepackage{booktabs}
\usepackage{multirow}
\usepackage{pifont}
\usepackage{marvosym}
\usepackage{colortbl}
\usepackage{xcolor}
\usepackage{listings}
\usepackage{lstlinebgrd}
\usepackage{algorithm}
\usepackage{placeins}
\usepackage[most]{tcolorbox}
\usepackage{hyperref}
\definecolor{gray}{HTML}{F2F2F2}
\newtcolorbox{graybox}{
    colback=gray,       
    colframe=teal!60!black,    
    arc=5pt,                   
    outer arc=5pt,             
    boxrule=0.8pt,             
    shadow={1.5pt}{-1.5pt}{0pt}{black!20!white}, 
    left=5pt,                 
    right=5pt,                
    top=4pt,                   
    bottom=4pt,                
    enhanced,                  
}

\title{NowcastDiT: Diffusion Transformers \\
are Effective Precipitation Nowcasters}

\author{Haoran Xu$^1$\thanks{Equal contribution. \Letter{} Corresponding author.}, Xingzhuo Guo$^1$\footnotemark[1], Yuchen Zhang$^2$, Jincheng Zhong$^3$, \\
\textbf{Jianmin Wang$^1$, Mingsheng Long$^{1,\text{\Letter}}$} \\
$^1$School of Software, BNRist, Tsinghua University, China \\
$^2$Envision Energy, China \\
$^3$Kling Team, Kuaishou Technology, China \\
\texttt{\{xuhaoran26,gxz23\}@mails.tsinghua.edu.cn} \\
\texttt{mingsheng@tsinghua.edu.cn}
}

\newif\ifshowtodos
\showtodostrue

\iclrfinalcopy 
\begin{document}

\maketitle
\lhead{Preprint}

\begin{abstract}
Precipitation nowcasting demands accurate short-term forecasts under strong spatiotemporal variability.
Diffusion models are well suited to modeling complex precipitation distributions, yet existing approaches often introduce increasingly specialized designs, leaving the capability of a standard diffusion architecture underexplored.
We show that a \emph{standard} Diffusion Transformer already provides a simple and scalable foundation for precipitation nowcasting, with domain-specific requirements accommodated naturally within its design space.
Based on this principle, we develop NowcastDiT and instantiate this flexibility through two complementary adaptations: a dynamics-aware noise prior for temporally coherent forecasts, and end-to-end reinforcement learning with timestep-aware rewards for meteorological skill.
Experiments on SEVIR and MRMS benchmarks show that NowcastDiT achieves state-of-the-art performance in both perceptual quality and meteorological skill.
These results suggest that standard DiT can serve as an effective foundation for precipitation nowcasting.
\end{abstract}

\section{Introduction}

Precipitation nowcasting aims to predict the short-term evolution of rainfall fields at high spatial and temporal resolutions.
An effective forecasting system is expected to accurately capture complex spatiotemporal dynamics, account for intrinsic uncertainty, and maintain reliability for rare but high-impact precipitation events~\citep{wmo2017guidelines}. From a generative modeling perspective, precipitation nowcasting is fundamentally a conditional generation task, with diffusion models~\citep{DDPMho2020denoising,SDE-song2020score,lipman2022flow} providing a natural framework given their strong capability for probabilistic image and video generation\citep{DiT-peebles2023scalable,wan2025wan,seedance2026seedance,agarwal2026cosmos}.

Beyond these general modeling capabilities, existing works have focused on several precipitation-specific requirements, particularly temporal coherence across forecast frames and meteorological skill in capturing high-impact events. To address them, existing generative methods introduce various task-specific designs
including physics-based motion constraints~\citep{zhang2023skilful}, intensity continuity regularization~\citep{gao2024prediff}, and decomposition-based modeling frameworks~\citep{yu2024diffcast, gong2024cascast, wen2026duocast}. Despite their different forms, these methods follow a common design philosophy: specializing the generative framework to explicitly encoding these requirements. While effective, this line of development narrows the exploration of foundational model designs for precipitation nowcasting, leaving a more fundamental question insufficiently explored: \emph{how necessary is task-specific specialization beyond a standard diffusion model?}

\begin{figure}[t]
    \vspace{-15pt}
    \centering
    \includegraphics[width=0.9\linewidth]{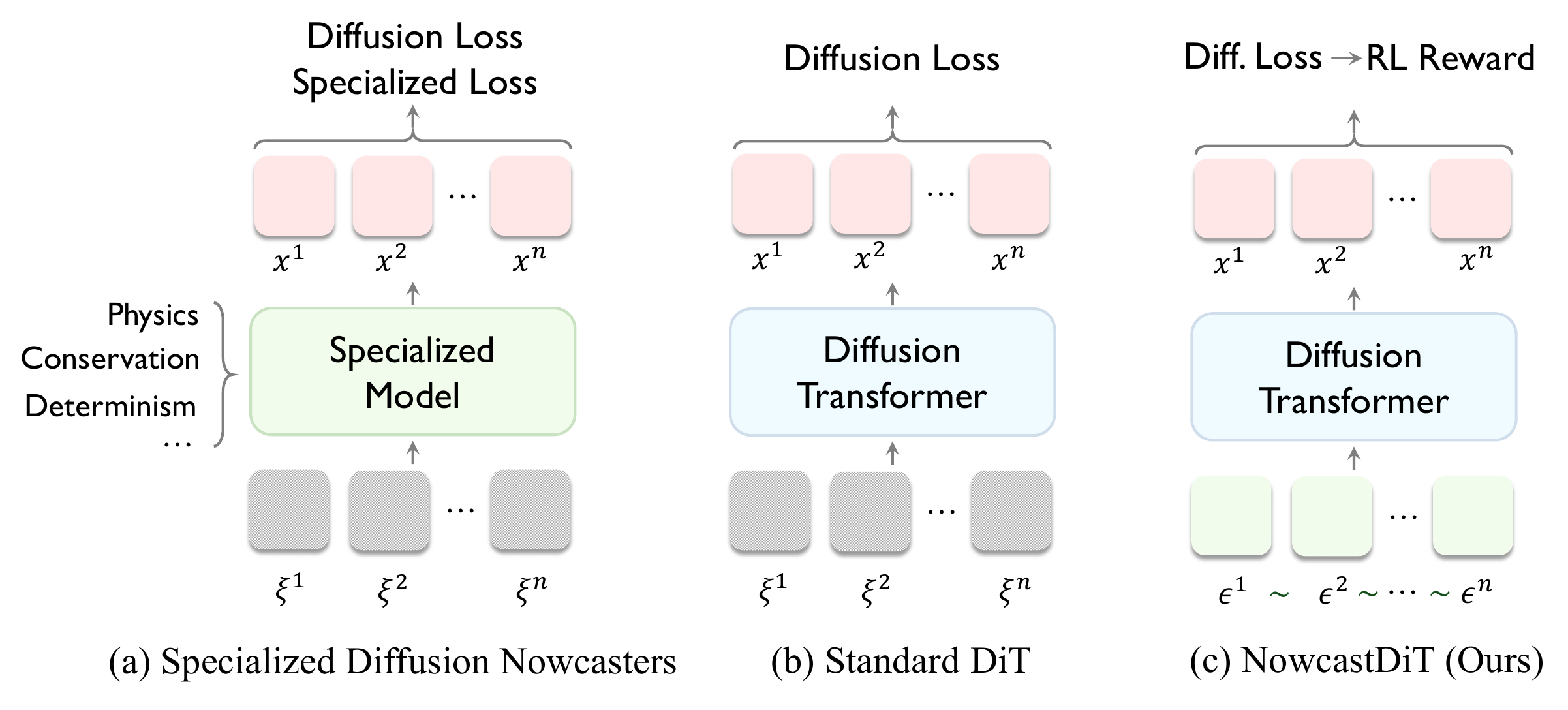}
    \vspace{-10pt}
    \caption{\textbf{Comparison of diffusion modeling paradigms for precipitation nowcasting.} \textbf{(a):} Specialized diffusion nowcasters encode domain knowledge through task-specific models or losses. \textbf{(b):} A standard DiT is applicable for precipitation nowcasting. \textbf{(c):} NowcastDiT adds minimal precipitation-specific adaptation towards a standard DiT without changing its backbone.}
    \vspace{-10pt}
    \label{fig:intro}
\end{figure}


In this paper, we argue that a standard Diffusion Transformer provides a sufficient foundation for precipitation nowcasting.
We posit that precipitation nowcasting is not fundamentally different from conditional video generation in terms of its core generative modeling requirements, as both require capturing complex spatiotemporal dynamics under uncertainty, capabilities that have already been demonstrated by modern video diffusion models~\citep{seedance2026seedance,agarwal2026cosmos}. Meanwhile, precipitation-specific requirements can be accommodated within the broader design space of diffusion models, as suggested by the versatility of modern diffusion models across diverse video generation tasks~\citep{guo2025dynamical,xue2025dancegrpo}


To this end, we propose the Nowcasting Diffusion Transformer (NowcastDiT), a simple and scalable framework 
built on a standard DiT with minimal precipitation-specific adaptations.
From the diffusion perspective, NowcastDiT introduces a dynamics-aware noise prior that captures shared evolution patterns and variations across forecast frames to enforce temporal coherence. From the training perspective, NowcastDiT incorporates end-to-end reinforcement learning with timestep-aware rewards to enhance meteorological skill.
On two widely recognized radar-based benchmarks, SEVIR and MRMS, NowcastDiT achieves state-of-the-art performance across both perceptual quality and meteorological skill metrics. These results suggest that standard DiT can serve as an effective foundation for precipitation nowcasting.

Our contributions are summarized as follows:

\begin{itemize}
    \item We revisit precipitation nowcasting from the perspective of standard diffusion models, and show that a standard DiT provides a sufficient foundation, with precipitation-specific requirements handled through targeted adaptation rather than model specialization.
    \item Based on this principle, we propose NowcastDiT, a simple and scalable framework that retains the standard DiT with a dynamics-aware noise prior to promote temporal coherence and reinforcement learning with timestep-aware rewards to enhance meteorological skill.
    \item Extensive experiments on SEVIR and MRMS demonstrate the state-of-the-art performance of NowcastDiT, validating standard DiT as an effective foundation for precipitation nowcasting.
\end{itemize}

\section{Preliminaries}
Precipitation nowcasting aims to predict $S$ future frames $\mathbf{x}^{1:S}$ from past frames $\mathbf{x}^{-S_0:0}$.
Following~\citet{rombach2022high}, we first train a VAE with encoder $E$ and decoder $D$, then generate $\mathbf{z}_1=E(\mathbf{x}^{S})$ conditioned on $\mathbf{c}=E(\mathbf{x}^{-S_0:0})$ within the latent space.
Diffusion models learn to denoise noise-corrupted data through simple linear interpolation, learning a conditional velocity field with the flow-matching objective:
\begin{align}
\mathbf{z}_t
=(1-t)\boldsymbol{\epsilon}+t\mathbf{z}_1,
\end{align}
\begin{align}
\mathcal{L}_{\mathrm{FM}}(\theta)
&=\mathbb{E}_{\mathbf{z}_1,\mathbf{c},\boldsymbol{\epsilon},t}
\left[\left\|
\mathbf{v}_{\theta}(\mathbf{z}_t,t\mid\mathbf{c})
-(\mathbf{z}_1-\boldsymbol{\epsilon})
\right\|_2^2\right],
\end{align}
where $t\in[0,1]$ denotes the diffusion timestep and $\boldsymbol{\epsilon}\sim\mathcal{N}(\mathbf{0},\mathbf{I})$ is Gaussian noise.
At inference, we solve $\frac{\mathrm{d}\mathbf{z}_t}{\mathrm{d}t}=\mathbf{v}_{\theta}(\mathbf{z}_t,t\mid\mathbf{c})$ from $\mathbf{z}_0=\boldsymbol{\epsilon}$ to $\hat{\mathbf{z}}_1$ and decode $\hat{\mathbf{x}}^{1:S}=D(\hat{\mathbf{z}}_1)$.

\section{Revisiting Precipitation Nowcasting Requirements}

\label{sec:revisit}

In this section, we revisit the modeling requirements of precipitation nowcasting and ask whether they necessitate specialized forecasting architectures. We organize the discussion around three aspects: spatiotemporal generative modeling, physical plausibility, and precipitation alignment.

\paragraph{Spatiotemporal generative modeling.}
Precipitation fields, like natural videos, are continuous spatiotemporal tensors whose structures span multiple scales, from localized convective cells to large-scale frontal systems. A suitable model must capture these hierarchical dependencies while representing the uncertainty of future evolution. The success of DiTs in video generation~\citep{sora2024, wan2025wan,seedance2026seedance} suggest that their scalable spatiotemporal representations provides a suitable foundation for this purpose.

\paragraph{Physical plausibility.}
Precipitation evolves through atmospheric processes, and plausible forecasts should exhibit consistent motion and physically reasonable temporal evolution. Explicit physical constraints offer one way to encourage such behavior, but they are not the only route. Modern generative models have demonstrated the ability to learn intuitive physics from video~\citep{ali2025world, ye2026world,agarwal2026cosmos}, suggesting that data-driven DiTs can also capture the coherent motion and local evolution from radar sequences.

\paragraph{Precipitation alignment.} Precipitation differs from natural video mainly in its data distribution and task objectives. Its evolution combines large-scale advection with localized growth and decay, with precipitation intensities commonly sparse and long-tailed. Moreover, generic generative objectives do not directly optimize meteorological skills such as Critical Success Index (CSI) and Heidke Skill Score (HSS). These discrepancies therefore call for precipitation-aware adaptation of the generative process and learning objective, but do not by themselves prescribe a specialized model or alter the underlying spatiotemporal modeling problem.



Existing nowcasting methods often couple such domain-specific mechanisms with specialized forecasting models or frameworks. This coupling obscures how much task-specific specialization is genuinely needed beyond the capabilities of a general-purpose generative model. The analysis above suggests a cleaner separation: retain a standard DiT for general spatiotemporal modeling, and introduce specialization only where precipitation departs from generic generation.

\begin{graybox}
     \textbf{Takeaway.} A standard DiT provides the core modeling capacity required for precipitation nowcasting, while precipitation-specific requirements can be addressed through targeted adaptations rather than model re-design.
\end{graybox}

\section{NowcastDiT}
\label{sec:method}

\begin{figure*}[t]
    \vspace{-15pt}
    \centering
    \includegraphics[width=\textwidth]{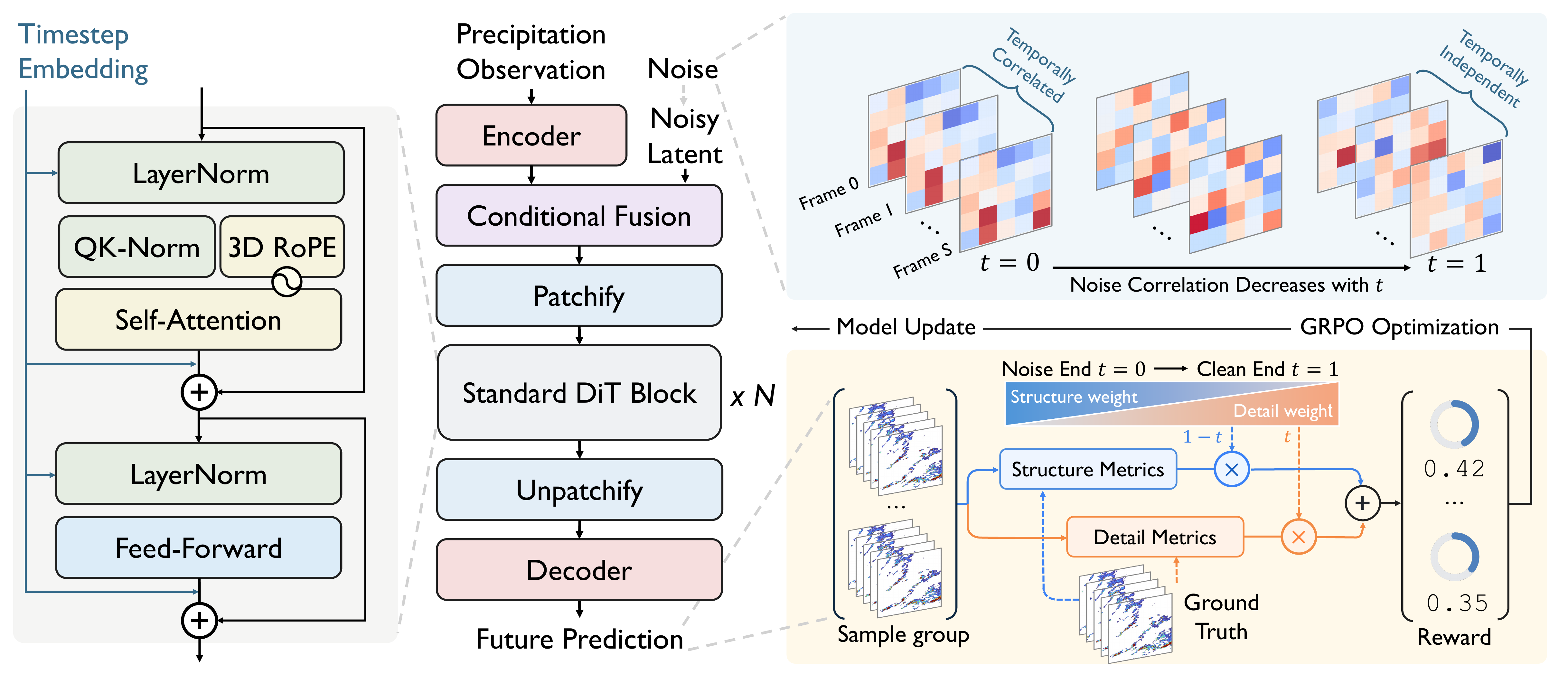}
    \vspace{-15pt}
    \caption{Overview of \textbf{NowcastDiT}. \textit{Left:} A standard Diffusion Transformer predicts future precipitation conditioned on encoded observations, with QK-Norm and 3D RoPE. \textit{Top right:} The dynamics-aware noise prior (DyPro) schedules cross-frame noise correlation from temporally correlated at the noise endpoint to independent at the clean endpoint. \textit{Bottom right:} Timestep-aware post-training shifts the reward emphasis from structure to detail along the denoising trajectory.}
    \label{fig:method_overview}
    \vspace{-5pt}
\end{figure*}

In this section, we present NowcastDiT, a precipitation nowcasting framework built on a standard Diffusion Transformer with minimal modifications. As illustrated in Figure~\ref{fig:method_overview}, it promotes temporal coherence through scheduled noise correlation and meteorological skill through timestep-aware rewards, framing precipitation nowcasting as a natural extension of video diffusion modeling.

\subsection{Transformer Backbone}

Precipitation nowcasting shares the core spatiotemporal modeling requirements of video generation, making a standard diffusion transformer a natural backbone. Following standard visual-conditioning designs in video diffusion models~\citep{blattmann2023stable,blattmann2023align,voleti2022mcvd,gupta2024walt}, NowcastDiT conditions future latent generation on historical observations by adding encoded observations to the noisy patch embeddings. One advantage of preserving a task-agnostic backbone is that NowcastDiT can inherit advances from broader generative modeling. Specifically, we incorporate two such advances that directly address the modeling requirements of precipitation nowcasting.

\textbf{QK-Norm.} Radar-based precipitation observations yield a sparse, long-tailed intensity distribution, which can produce large variations in token feature magnitudes and attention logits. To keep attention computation well scaled during optimization, we apply QK-Norm~\citep{qknorm-henry2020query} to normalize query and key vectors before their dot product.

\textbf{3D RoPE.} Precipitation sequences form a regular spatiotemporal grid, and their evolution is characterized by relative displacement across frames and spatial locations. Following CogVideoX~\citep{yang2024cogvideox} and Wan~\citep{wan2025wan}, we apply RoPE~\citep{su2024roformer} along the temporal, height, and width axes, enabling attention to encode relative offsets along all three dimensions in spacetime.

Overall, NowcastDiT retains a deliberately simple framework: it relies on a standard DiT without task-specialized forecasting modules, explicit physical constraints, or cascaded architectures.

\subsection{Temporal Coherence with Scheduled Noise Correlation}
\label{sec:dypro}


Standard video diffusion formulations typically adopt i.i.d. Gaussian noise over the spatiotemporal volume, yielding independent noise realizations across forecast frames. While simple and broadly applicable, this frame-wise independence is less aligned with the strong temporal coherence inherent in precipitation evolution.

\textbf{Dynamics-aware noise prior.} 
Existing approaches that adapt pretrained image diffusion models for video generation have shown that correlating noise across adjacent forecast frames can improve temporal consistency~\citep{ge2023preserve,chang2024warped}. 
Motivated by this insight, we introduce a dynamics-aware noise prior \textit{(DyPro)} that correlates noise across forecast frames. To construct this prior, we draw on the progressive noise schedule following Dynamical Diffusion~\citep{guo2025dynamical}, with the strength of cross-frame correlation controlled by diffusion timestep.
Let $s\in\{1,\ldots,S\}$ index the forecast frames, and let $\boldsymbol{\epsilon}^{1:S}_t$ denote the corresponding noise sequence at timestep $t$. For a fixed diffusion timestep $t$, DyPro recursively constructs
\begin{equation}
\label{eq:dypro_noise_def}
\boldsymbol{\epsilon}^1_t = \boldsymbol{\xi}^1,\quad\boldsymbol{\epsilon}^s_t = \sqrt{\gamma_t}\boldsymbol{\epsilon}^{s-1}_t
+ \sqrt{1 - \gamma_t}\,\boldsymbol{\xi}^s .
\end{equation}
Here, $\boldsymbol{\xi}^s \sim \mathcal{N}(\mathbf{0},\mathbf{I})$ are independent Gaussian samples, and the diffusion-timestep-dependent inheritance weight is
\begin{equation}
\gamma_t =
\frac{\alpha^2(1-t)^2}{1+\alpha^2(1-t)^2},
\end{equation}
where hyperparameter $\alpha \geq 0$ controls the temporal dependence together with $t$. The recursion preserves the standard Gaussian marginal of each frame, $\boldsymbol{\epsilon}^s_t \sim \mathcal{N}(\mathbf{0},\mathbf{I})$, while modifying only the cross-frame covariance. Specifically, for two frames $k$ steps apart, $\operatorname{Cov}(\boldsymbol{\epsilon}^{s}_t,\boldsymbol{\epsilon}^{s-k}_t)=\gamma_t^{k/2}\mathbf{I}$. Thus, adjacent frames share the most similar noise, and this similarity decays geometrically with temporal distance $k$. As denoising proceeds, $\gamma_t$ decreases from $\alpha^2/(1+\alpha^2)$ at the beginning to zero at the clean endpoint. DyPro therefore promotes spatiotemporal coherence over the fixed radar grid during early denoising, while progressively relaxing the coupling during later stages to preserve variations among individual forecast frames associated with localized precipitation growth, decay, and intensity evolution.

\begingroup
\definecolor{codeblue}{rgb}{0.25,0.5,0.5}
\definecolor{codekw}{rgb}{0.85, 0.18, 0.50}

\definecolor{codesign}{RGB}{0, 0, 255}
\definecolor{codefunc}{rgb}{0.85, 0.18, 0.50}

\definecolor{highlightcolor}{HTML}{F2F2F2}

\lstdefinelanguage{PythonFuncColor}{
  language=Python,
  keywordstyle=\color{blue}\bfseries,
  commentstyle=\color{codeblue},
  stringstyle=\color{orange},
  showstringspaces=false,
  basicstyle=\ttfamily\small,
  literate=
    {*}{{\color{codesign}*}}{1}
    {-}{{\color{codesign}-}}{1}
    {+}{{\color{codesign}+}}{1}
    {/}{{\color{codesign}/}}{1}
    {dataloader}{{\color{codefunc}dataloader}}{1}
    {sample_t}{{\color{codefunc}sample\_t}}{1}
    {randn}{{\color{codefunc}randn}}{1}
    {randn_like}{{\color{codefunc}randn\_like}}{1}
    {correlate}{{\color{codefunc}correlate}}{1}
    {decorrelate}{{\color{codefunc}decorrelate}}{1}
    {jvp}{{\color{codefunc}jvp}}{1}
    {stopgrad}{{\color{codefunc}stopgrad}}{1}
    {l2_loss}{{\color{codefunc}l2\_loss}}{1}
    {net}{{\color{codefunc}net}}{1},
  escapeinside={(*}{*)},
}

\lstset{
  language=PythonFuncColor,
  backgroundcolor=\color{white},
  basicstyle=\fontsize{8pt}{8.4pt}\ttfamily\selectfont,
  columns=fullflexible,
  keepspaces=true,
  breaklines=true,
  captionpos=b,
  aboveskip=4pt,
  belowskip=4pt,
}

\begin{figure}[t]
    \vspace{-15pt}
    \centering
    \begin{minipage}[t]{0.485\linewidth}
        \vspace{0pt}
        \begin{algorithm}[H]
            \caption{DyPro Training}
            \label{alg:code_train}
\begin{lstlisting}[linebackgroundcolor={\ifnum\value{lstnumber}=8 \color{highlightcolor}\fi}]
# x: training latents; c: history
# net(z, t, c): predicts x - e

t = sample_t()

# Construct time-dependent noise
e_init = randn_like(x)
e = correlate(e_init, t)

z = t * x + (1 - t) * e
v = x - e

v_pred = net(z, t, c)
loss = l2_loss(v - v_pred)
update(net, loss)
\end{lstlisting}
        \end{algorithm}
    \end{minipage}\hfill
    \begin{minipage}[t]{0.485\linewidth}
        \vspace{0pt}
        \begin{algorithm}[H]
            \caption{DyPro Sampling (w/o CFG)}
            \label{alg:code_sample}
\begin{lstlisting}[linebackgroundcolor={\ifnum\value{lstnumber}=5 \color{highlightcolor}\fi\ifnum\value{lstnumber}=11 \color{highlightcolor}\fi\ifnum\value{lstnumber}=12 \color{highlightcolor}\fi}]
# c: history; shape: latent shape
# N: number of sampling steps

ts = linspace(0, 1, N + 1)
z = correlate(randn(shape), 0)

for t, t_next in zip(ts[:-1], ts[1:]):
    v = net(z, t, c)
    x = z + (1 - t) * v
    e = z - t * v
    e_init = decorrelate(e, t)
    e = correlate(e_init, t_next)
    z = t_next * x + (1 - t_next) * e

return z
\end{lstlisting}
        \end{algorithm}
    \end{minipage}
\end{figure}
\endgroup

\textbf{Training and Inference.} DyPro is independent of the backbone, preserving the model's input and output shapes and the form of the training objective. During training, we use DyPro noise to construct $\mathbf{z}_t$ and the flow-matching target (Algorithm~\ref{alg:code_train}). At inference, we use a sampler following Dynamical Diffusion that updates noise correlation at each step (Algorithm~\ref{alg:code_sample}). The sampler requires one network evaluation per step, matching standard Euler sampling in styles.

\subsection{Meteorological Skill with Timestep-Aware Rewards}
\label{sec:post-training}


Standard generative models are typically optimized with surrogate objectives such as flow matching and evaluated by metrics aligned with generative fidelity. Yet precipitation nowcasting also emphasizes task-specific meteorological skill that does not necessarily align with these objectives, making flow matching alone insufficient for optimizing forecasting performance.

Existing work has shown that task-specific evaluation metrics can serve directly as verifiable rewards for optimizing generative prediction models~\citep{wu2025rlvr,xue2025dancegrpo}. 
Inspired by the pretraining--posttraining paradigm of foundation models, we adopt a two-stage training strategy that separates general generative modeling from meteorological skill optimization:
\begin{itemize}
\item \textbf{Stage 1:} Pretrain a diffusion model from scratch with surrogate flow
matching objective.
\item \textbf{Stage 2:} Posttrain the diffusion model by directly optimizing
meteorological skills.
\end{itemize}

\textbf{Skill optimization via GRPO.} We formulate the posttraining stage as a GRPO objective for diffusion models~\citep{shao2024deepseekmath,liu2025flowgrpo}. For each conditioning context paired with a ground-truth future sequence
$\mathbf{x}^*$, we sample a group of forecasts $\{\hat{\mathbf{x}}^{(1)}, \hat{\mathbf{x}}^{(2)}, \ldots, \hat{\mathbf{x}}^{(G)}\}$ through a sampling trajectory where selected ODE transitions are converted into SDE transitions that serve as stochastic policy actions with tractable likelihoods, following the design of MixGRPO~\citep{li2025mixgrpo}.
We score each $\hat{\mathbf{x}}^{(i)}$ against $\mathbf{x}^*$ and update the stochastic transitions using the clipped group-relative objective
\begin{align}
\hat{A}_n^i
&=
\frac{
R(\hat{\mathbf{x}}^{(i)},\mathbf{x}^*;t_n)
-
\operatorname*{mean}
\left(
\left\{R(\hat{\mathbf{x}}^{(j)},\mathbf{x}^*;t_n)\right\}_{j=1}^{G}
\right)}
{
\operatorname{std}
\left(
\left\{R(\hat{\mathbf{x}}^{(j)},\mathbf{x}^*;t_n)\right\}_{j=1}^{G}
\right)},
\label{eq:main_group_advantage}
\end{align}
\begin{align}
\mathcal{J}_{\mathrm{GRPO}}(\theta)
=\frac{1}{G|\mathcal{W}_{\ell}|}
\sum_{n\in\mathcal{W}_{\ell}}\sum_{i=1}^{G}
\min\Big(\rho_n^i(\theta)\hat{A}_n^i,\operatorname{clip}\!\left(\rho_n^i(\theta),
1-\varepsilon,1+\varepsilon\right)\hat{A}_n^i\Big),
\label{eq:main_grpo_objective}
\end{align}
where $R$ denotes the reward evaluated on the final forecast, $\mathcal{W}_{\ell}$ indexes the stochastic transitions included in the GRPO update, and $\rho_n^i$ is the policy likelihood ratio for forecast $i$ at transition $n$. Further details on GRPO optimization and mixed ODE--SDE sampling are provided in Appendix~\ref{app:grpo_math} and Appendix~\ref{app:mixed_sampling}, respectively.


\textbf{Timestep-aware rewards.}
Reinforcement learning for diffusion models typically uses the same terminal reward across all denoising transitions. However, different denoising transitions are expected to contribute differently to the final forecast: transitions near the noise endpoint establish precipitation coverage and large-scale spatial patterns, whereas transitions near the clean endpoint refine local intensity variations and heavy-precipitation cores. This uniform supervision may therefore be suboptimal. We introduce a timestep-aware terminal reward that gradually shifts its emphasis from precipitation structure to local detail as denoising proceeds:
\begin{align}
R_{\mathrm{noise}}(\hat{\mathbf{x}},\mathbf{x}^*)
&= \sum_{\tau\in\mathcal{T}_{\mathrm{low}}}\lambda_{\tau}\mathrm{CSI}^{\mathrm{pool}}_{\tau}(\hat{\mathbf{x}},\mathbf{x}^*)+\lambda_s\mathrm{SSIM}(\hat{\mathbf{x}},\mathbf{x}^*),
\label{eq:noise_reward}\\
R_{\mathrm{clean}}(\hat{\mathbf{x}},\mathbf{x}^*)
&= \sum_{\tau\in\mathcal{T}_{\mathrm{high}}}\lambda_{\tau}\mathrm{CSI}_{\tau}(\hat{\mathbf{x}},\mathbf{x}^*)-\lambda_p\mathrm{LPIPS}(\hat{\mathbf{x}},\mathbf{x}^*),
\label{eq:clean_reward}\\
R(\hat{\mathbf{x}},\mathbf{x}^*;t)
&= (1-t)R_{\mathrm{noise}}(\hat{\mathbf{x}},\mathbf{x}^*)+tR_{\mathrm{clean}}(\hat{\mathbf{x}},\mathbf{x}^*),
\label{eq:stage_aware_reward}
\end{align}
where $\mathcal{T}_{\text{low}}$ and $\mathcal{T}_{\text{high}}$ denote the light-precipitation and heavy-precipitation threshold sets, respectively. The coefficients $\lambda_s$ and $\lambda_p$ balance the constituent metrics, while the remaining weight is distributed uniformly across the corresponding CSI thresholds:
$\lambda_{\tau}=(1-\lambda_s)/|\mathcal{T}_{\text{low}}|$ for $R_{\text{noise}}$ and
$\lambda_{\tau}=(1-\lambda_p)/|\mathcal{T}_{\text{high}}|$ for $R_{\text{clean}}$.
We empirically set $\lambda_s=0.5$ and $\lambda_p=0.9$.

Specifically, $R_{\mathrm{noise}}$ combines spatially tolerant low-threshold pooled CSI for event coverage with SSIM~\citep{wang2004SSIM} for local structural agreement, while $R_{\mathrm{clean}}$ combines high-threshold CSI for localized intense precipitation with negative LPIPS~\citep{zhang2018unreasonable} for perceptual similarity in deep feature space. Since $t=0$ and $t=1$ correspond to the noise and clean endpoints, respectively, the reward gradually shifts from $R_{\mathrm{noise}}$ to $R_{\mathrm{clean}}$, emphasizing precipitation structure early and local detail later. All rewards are computed on the final decoded forecast, with the active transition time determining the metric mixture used for credit assignment in Equation~\ref{eq:main_group_advantage}.

\section{Experiments}
\begin{table}[t]
\centering
\vspace{-15pt}
\caption{
Performance comparison on SEVIR and MRMS.
$\uparrow$ indicates higher is better and $\downarrow$ indicates lower is better.
Bold and underlined values denote the best and second-best results, respectively.}
\label{tab:main}
\fontsize{8}{10}\selectfont
\setlength{\tabcolsep}{0.9pt}
\renewcommand{\arraystretch}{1.05}
\begin{tabular*}{\linewidth}{@{\extracolsep{\fill}}l*{6}{c}@{\hspace{8pt}}l*{6}{c}@{}}
\toprule
\multicolumn{7}{c}{\textbf{SEVIR}} & \multicolumn{7}{c}{\textbf{MRMS}} \\
\addlinespace[2pt]
Method & CSI$^{\uparrow}$ & CSI$_{181}^{\uparrow}$ & CSI$_{219}^{\uparrow}$ & HSS$^{\uparrow}$ & LPIPS$^{\downarrow}$ & SSIM$^{\uparrow}$
    & Method & CSI$^{\uparrow}$ & CSI$_{16}^{\uparrow}$ & CSI$_{32}^{\uparrow}$ & HSS$^{\uparrow}$ & LPIPS$^{\downarrow}$ & SSIM$^{\uparrow}$ \\
\midrule
ConvLSTM & 0.2912 & 0.0684 & 0.0336 & 0.3571 & 0.2989 & \underline{0.7257}
    & ConvLSTM & 0.2159 & 0.0514 & 0.0172 & 0.2877 & 0.3013 & 0.8978 \\
PhyDNet & 0.2874 & 0.0664 & 0.0211 & 0.3523 & 0.3081 & 0.7252
    & PhyDNet & 0.2332 & 0.0635 & 0.0268 & 0.3111 & 0.3005 & \underline{0.9001} \\
Earthformer & 0.2752 & 0.0496 & 0.0213 & 0.3391 & 0.3376 & 0.7142
    & Earthformer & 0.2353 & 0.0691 & 0.0297 & 0.3152 & 0.3001 & 0.8994 \\
SimVP & 0.2951 & 0.0753 & 0.0416 & 0.3630 & 0.3113 & 0.7247
    & SimVP & \underline{0.2366} & 0.0657 & 0.0277 & 0.3155 & 0.2992 & \textbf{0.9006} \\
AlphaPre & \underline{0.2980} & 0.0860 & 0.0436 & 0.3677 & 0.2897 & \textbf{0.7301}
    & AlphaPre & 0.2309 & 0.0587 & 0.0211 & 0.2933 & 0.3014 & 0.8944 \\
\midrule
DiffCast & 0.2977 & 0.0915 & \underline{0.0571} & \underline{0.4033} & 0.1812 & 0.6774
    & NowcastNet & 0.2244 & \underline{0.1145} & \underline{0.0729} & 0.3083 & 0.2007 & 0.8502 \\
PreDiff & 0.2859 & 0.0893 & 0.0487 & 0.3647 & \underline{0.1543} & 0.7006
    & PreDiff & 0.2283 & 0.1083 & 0.0695 & 0.3148 & \underline{0.1906} & 0.8633 \\
CasCast & 0.2905 & 0.0930 & 0.0503 & 0.3703 & 0.1583 & 0.7012
    & CasCast & 0.2224 & 0.0925 & 0.0527 & 0.3064 & 0.2045 & 0.8791 \\
\midrule
DiT & 0.2926 & \underline{0.0948} & 0.0555 & 0.3733 & 0.1545 & 0.7048
    & DiT & 0.2343 & 0.1015 & 0.0611 & \underline{0.3223} & 0.1953 & 0.8824 \\
NowcastDiT & \textbf{0.3240} & \textbf{0.1241} & \textbf{0.0750} & \textbf{0.4148} & \textbf{0.1469} & 0.7197
    & NowcastDiT & \textbf{0.2696} & \textbf{0.1373} & \textbf{0.0958} & \textbf{0.3668} & \textbf{0.1827} & 0.8642 \\
\bottomrule
\end{tabular*}
\vspace{-15pt}
\end{table}

We evaluate NowcastDiT on two radar-based benchmarks, SEVIR and MRMS. Comparisons with deterministic and generative nowcasting baselines assess both meteorological skill and perceptual quality. Additional experiments examine how our design choices affect forecasting performance.

\subsection{Experimental Setup}
\label{sec:experimental_setup}

\begin{figure}[t]
    \centering
    \vspace{-15pt}
    \includegraphics[width=\linewidth]{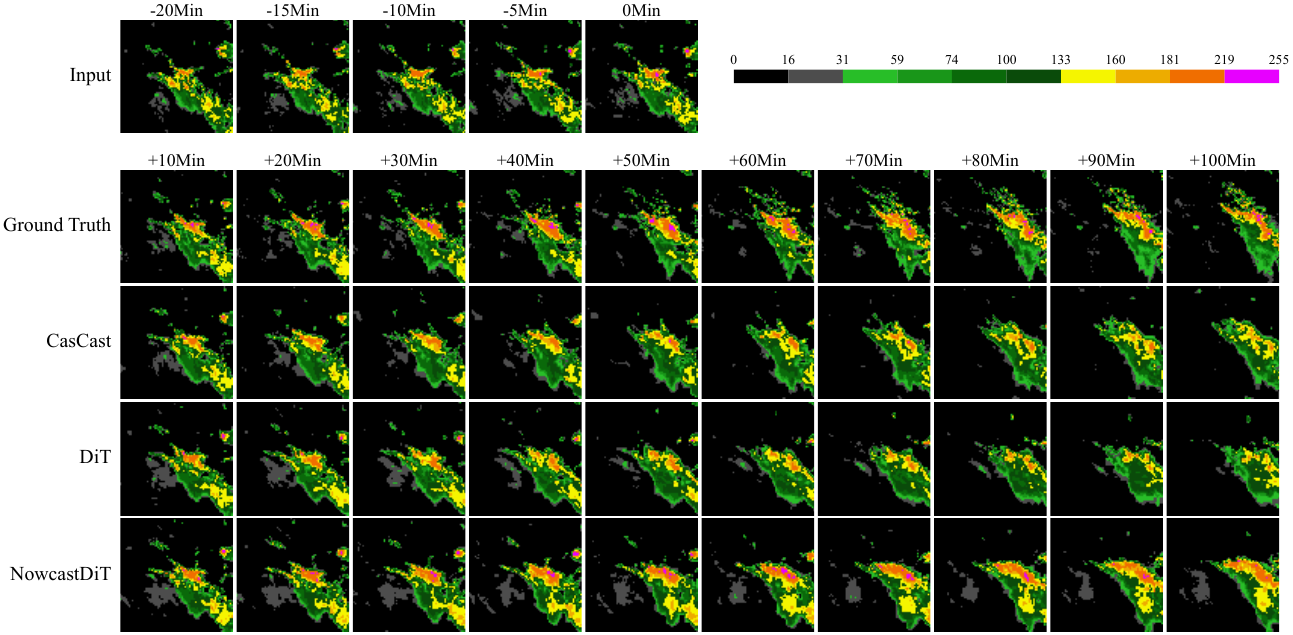}
    \vspace{-15pt}
    \caption{\textbf{Qualitative comparison on SEVIR.} Forecasts are shown every 10 minutes up to 100 minutes. Colors indicate VIL values.}
    \vspace{-15pt}
    \label{fig:sevir_qualitative}
\end{figure}

\textbf{Datasets.} SEVIR~\citep{veillette2020sevir} is a spatiotemporal radar observation dataset covering weather events across the United States. Following prior work~\citep{yu2024diffcast,lin2025alphapre}, we use Vertically Integrated Liquid (VIL) observations at a 5-minute temporal resolution to predict 20 future frames given 5 observed frames at $128 \times 128$ resolution. MRMS~\citep{zhang2016mrms} is a composite radar dataset collected over the contiguous United States. Following NowcastNet~\citep{zhang2023skilful}, we use precipitation-rate observations at a 10-minute temporal resolution to predict 20 future frames given 4 observed frames at $256 \times 256$ resolution. Additional dataset details, splits, and preprocessing are provided in Appendix~\ref{app:data_details}.


\textbf{Evaluation.} We compare NowcastDiT with deterministic and generative methods. The deterministic baselines include ConvLSTM~\citep{shi2015convolutional}, PhyDNet~\citep{guen2020disentangling}, Earthformer~\citep{gao2022earthformer}, SimVP~\citep{gao2022simvp}, and AlphaPre~\citep{lin2025alphapre}, while the generative baselines include NowcastNet~\citep{zhang2023skilful}, PreDiff~\citep{gao2024prediff}, DiffCast~\citep{yu2024diffcast}, and CasCast~\citep{gong2024cascast}. We evaluate meteorological skill using the Critical Success Index (CSI)~\citep{schaefer1990critical} and Heidke Skill Score (HSS)~\citep{jolliffe2012forecast}, and perceptual quality using the Structural Similarity Index (SSIM)~\citep{wang2004SSIM} and Learned Perceptual Image Patch Similarity (LPIPS)~\citep{zhang2018unreasonable}. CSI measures event detection accuracy, while HSS accounts for agreement expected by chance. Both are computed from pixelwise contingency counts aggregated over the evaluation set at each forecast lead time. We report their means over lead times and thresholds, using $\{16,74,133,160,181,219\}$ in the VIL for SEVIR and $\{1,2,4,8,16,32\}$~mm\,h$^{-1}$ for MRMS. We additionally report high-threshold CSI to assess intense precipitation events. SSIM measures local structural agreement, whereas LPIPS measures perceptual distance in deep feature space. Higher CSI, HSS, and SSIM and lower LPIPS indicate better performance. Metric definitions are provided in Appendix~\ref{app:evaluation_metrics}.

\textbf{Implementation Details.} We train NowcastDiT using AdamW with $(\beta_1,\beta_2)=(0.9,0.95)$, a batch size of 32, and learning rates of $10^{-4}$ for pretraining and $10^{-5}$ for post-training. The default backbone has 12 Transformer blocks, a hidden dimension of 768, and 12 attention heads. The DyPro parameter $\alpha$ is set to 0.5. Post-training uses a group size of 16, 10 denoising steps, an SDE window size of 4, and a policy ratio clipping range of $10^{-4}$. For reward computation, we apply $4\times4$ max pooling to low-threshold CSI and set the SSIM and LPIPS mixing weights to 0.5 and 0.9, respectively. All training runs use four NVIDIA A100 GPUs with 80 GB of memory each. Additional post-training settings are provided in Appendix~\ref{app:rlvr_impl}.

\subsection{Main Results}
We compare NowcastDiT with deterministic and generative nowcasting methods on SEVIR and MRMS, reported in Table~\ref{tab:main}. The standard DiT baseline already achieves competitive performance against specialized nowcasting models, demonstrating the strength of its general-purpose spatiotemporal modeling. Relative to the strongest baselines, NowcastDiT improves aggregate CSI by 8.7\% and 15.1\%, and HSS by 2.9\% and 13.8\%, on SEVIR and MRMS, respectively. It also reduces LPIPS by 4.8\% and 4.1\% while achieving the best SSIM among generative methods on SEVIR. On high-threshold CSI that directly reflects the ability to detect rare and operationally important intense rainfall events, NowcastDiT achieves relative gains of 19.9\%--31.4\% across the reported high thresholds, showing a clear advantage in heavy-rainfall prediction. We also provide per-lead-time CSI and HSS curves, shown in Appendix~\ref{app:lead_time_results}. Taken together, these results show that a standard DiT, with precipitation-specific alignment can achieve strong nowcasting performance while retaining a simple architecture.

\begin{figure}[t]
    \centering
    \vspace{-15pt}
    \includegraphics[width=\linewidth]{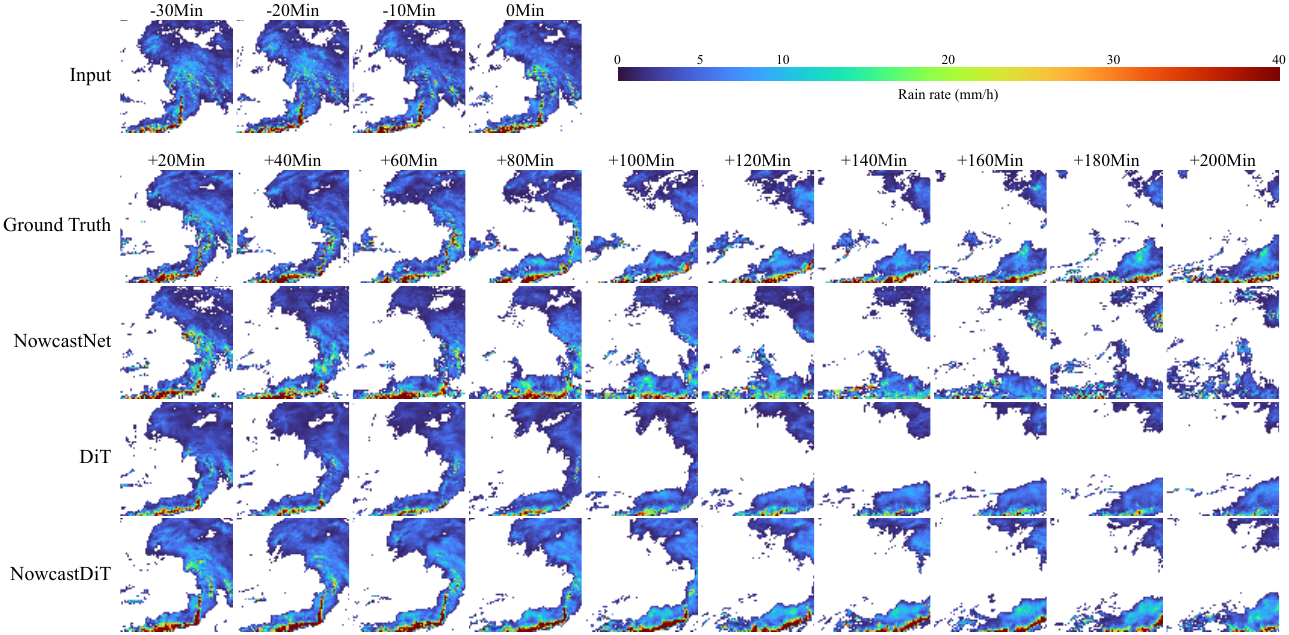}
    \vspace{-15pt}
    \caption{\textbf{Qualitative comparison on MRMS.} Forecasts are shown every 20 minutes up to 200 minutes. Colors indicate precipitation rate in mm\,h$^{-1}$.}
    \vspace{-10pt}
    \label{fig:mrms_qualitative}
\end{figure}

\textbf{Qualitative comparisons.} Figures~\ref{fig:sevir_qualitative} and~\ref{fig:mrms_qualitative} compare forecast sequences on SEVIR and MRMS, respectively. In the SEVIR example, NowcastDiT retains localized high-intensity regions at longer lead times, whereas the DiT baseline produces smoother precipitation fields. In the MRMS example, NowcastDiT better preserves the shape of the main rainband. These selected examples complement the test-set comparisons in Table~\ref{tab:main}.

\subsection{Analyses}

We analyze how the following components and design choices affect model performance on SEVIR.

\begin{figure*}[t]
    \vspace{-15pt}
    \centering
    \includegraphics[width=\textwidth]{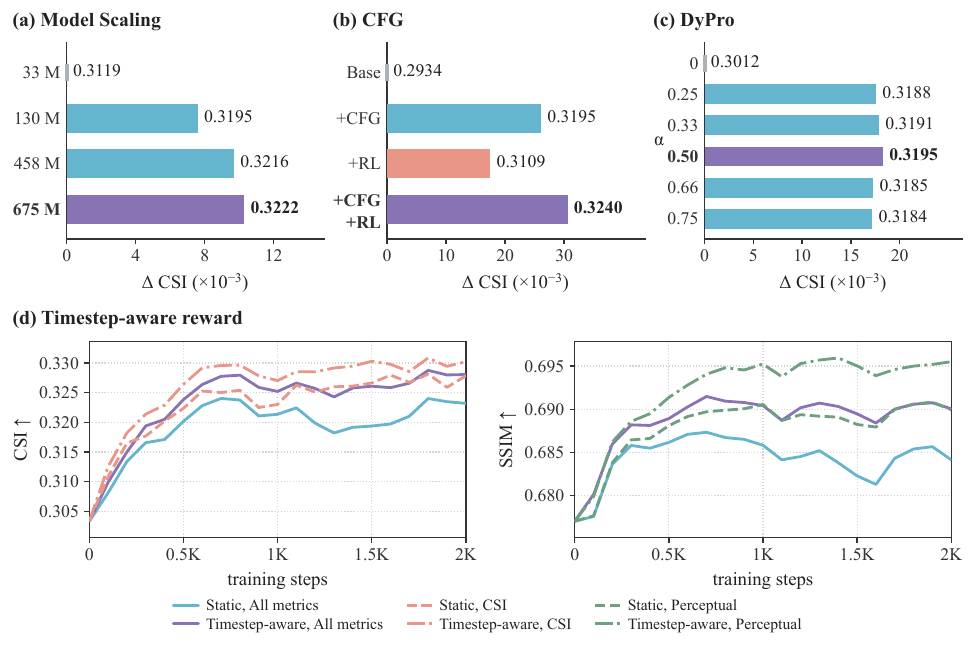}
    \vspace{-20pt}
    \caption{\textbf{Analysis on SEVIR.} Bar lengths indicate CSI gains relative to the first setting in each panel;
labels report absolute CSI values.}
    \label{fig:analysis_4panel}
    \vspace{-15pt}
\end{figure*}

\textbf{Model parameter scaling.} 
Diffusion Transformer is validated as a scalable backbone, and we investigate whether this scaling benefits extend to precipitation nowcasting. 
Following~\citet{DiT-peebles2023scalable}, we evaluate NowcastDiT across a range of model sizes. 
As shown in Figure~\ref{fig:analysis_4panel}(a), increasing model size consistently improves aggregate CSI. 
This finding supports model scaling as an effective strategy for strengthening the forecasting capabilities of a standard DiT backbone, which forms the initial motivation of NowcastDiT.

\textbf{CFG and skill-aware post-training.}
Classifier-free guidance (CFG)~\citep{cfg-ho2022classifier} is a common capability of standard diffusion models and should be naturally inherited when adopting them for precipitation nowcasting. We therefore examine whether skill-aware post-training can provide complementary gains on top of CFG, whose integration with diffusion post-training is not always trivial~\citep{zheng2026diffusionnft}. As shown in Figure~\ref{fig:analysis_4panel}(b), both CFG and RL improve aggregate CSI over the base model, and their combination achieves the best performance. This suggests that meteorological alignment provides an additional optimization axis beyond standard diffusion guidance. Detailed methods and results are provided in Appendix~\ref{app:component_ablation}.


\textbf{DyPro correlation strength.} For simplicity, NowcastDiT adopts the factor $\alpha=0.5$ controlling temporal correlations in DyPro. To assess the senstivity, 
we vary the DyPro correlation strength $\alpha\in\{0, 0.25, 0.33, 0.66, 0.75\}$, where $\alpha=0$ infers the i.i.d. noise baseline. 
Figure~\ref{fig:analysis_4panel}(c) shows that all tested nonzero strengths improve aggregate CSI, with performance peaking at a moderate strength and declining as the correlation increases further, indicating the robustness of $\alpha$ in DyPro.
Detailed results are provided in Appendix~\ref{app:dypro_results}.


\textbf{Timestep-aware reward.} The reinforcement learning stage of NowcastDiT uses a stage-aware reward to account for the different roles of reward metrics near the noise and clean endpoints. We compare it with a static baseline that mixes the same rewards without timestep-dependent weighting. As shown in Figure~\ref{fig:analysis_4panel}(d), timestep-aware weighting consistently performs better. To isolate this effect, we further repeat the comparison under CSI-only and perceptual-only objectives (i.e., $\lambda_s=\lambda_p=0$ and $\lambda_\tau=0$, respectively), where the same trend holds. These results confirm the benefit of timestep-aware weighting across reward objectives, while mixed metrics provide a better balance between meteorological skill and perceptual quality.

\section{Related Work}

\paragraph{Diffusion models for precipitation nowcasting.} 
Diffusion models have become an important paradigm for probabilistic nowcasting due to their stable training dynamics and strong generative fidelity~\citep{gao2024prediff,yu2024diffcast}.
Existing diffusion-based approaches often incorporate task-specific mechanisms to improve meteorological applicability~\citep{gao2024prediff,yu2024diffcast,gong2024cascast,wen2026duocast}. 
PreDiff~\citep{gao2024prediff} introduces intensity continuity guidance over a spatiotemporal architecture to strength physical alignment. 
DiffCast~\citep{yu2024diffcast} introduces decomposited training phases for the modeling of global deterministic motion and local stocastic variations. 
CasCast~\citep{gong2024cascast} employs a cascaded DiT framework in latent space to decouple deterministic and stochastic precipitation modeling for improved efficiency and extreme event fidelity.
These approaches couple domain-specific mechanisms with architectural and framework design, leaving unresolved whether specialized designs are necessary for effective precipitation nowcasting.
In contrast, NowcastDiT examines how a standard DiT can support effective nowcasting and address domain-specific requirements within its design space.

\paragraph{Advances in generative modeling techniques.}
Recent advances in image and video generation have substantially broadened the modeling capabilities of diffusion transformers~\citep{DiT-peebles2023scalable,SD3-esser2024scaling,ma2025latte,wan2025wan}.
Architectural refinements improve attention stability and positional modeling~\citep{qknorm-henry2020query,su2024roformer,gupta2024walt,yang2024cogvideox}.
Noise-prior designs introduce temporal dependencies to promote coherent video generation~\citep{ge2023preserve,chang2024warped,guo2025dynamical}.
Post-training strategies align generated outputs with task-specific objectives through reinforcement learning~\citep{black2024training,liu2025flowgrpo,xue2025dancegrpo,li2025mixgrpo,wu2025rlvr}.
Together, these developments establish the feasibility of meeting domain-specific modeling requirements within the design space of a standard DiT.
Building on this foundation, NowcastDiT further develops noise-prior and post-training designs to model precipitation dynamics and enhance meteorological skill.

\section{Conclusion}
This work demonstrates that a standard Diffusion Transformer, with targeted adaptations to its diffusion and training strategies, is sufficient for effective precipitation nowcasting. NowcastDiT combines DyPro for temporally coherent forecasting with skill-aware post-training for meteorological alignment. Experiments on SEVIR and MRMS show strong performance in both meteorological skill and perceptual quality, supporting standard DiTs as a simple and scalable framework for precipitation nowcasting. Future work will evaluate its generalization across observation systems, resolutions, and longer forecast horizons.

\clearpage
\subsection*{AI use statement}

AI tools assisted with drafting and revising portions of the manuscript, language polishing, literature search, figure preparation, verification of mathematical proof and use of the ICLR 2027 template. The authors personally reviewed and revised the AI-assisted content, verified its accuracy, and take full responsibility for the final manuscript, including all claims and results.

\bibliography{iclr2027_conference}
\bibliographystyle{iclr2027_conference}

\clearpage
\appendix
\raggedbottom
\newcommand{\appendixtablestyle}{%
    \centering
    \small
    \setlength{\tabcolsep}{6pt}%
    \renewcommand{\arraystretch}{1.1}%
    \setlength{\abovecaptionskip}{6pt}%
    \setlength{\belowcaptionskip}{6pt}%
}

\section{Model Architecture and Configuration}
\label{app:backbone_details}

\subsection{VAE Architecture and Training}
\label{app:vae_architecture}
We adopt a 2D VAE which compress only the spatial resolution which acts independently across frames. The pretrained VAE provides the encoder $E$ and decoder $D$ and remains frozen during flow-matching pretraining and RL post-training. Table~\ref{tab:vae_arch} summarizes its architecture and training settings.

\begin{table}[!ht]
\appendixtablestyle
\caption{VAE architecture and training configuration. Parameters include the encoder, decoder, and latent projections.}
\label{tab:vae_arch}
\begin{tabular}{@{}>{\raggedright\arraybackslash}p{0.36\linewidth}*{2}{>{\centering\arraybackslash}p{0.25\linewidth}}@{}}
\toprule
\textbf{Setting} & \textbf{SEVIR} & \textbf{MRMS} \\
\midrule
\rowcolor{black!10}\multicolumn{3}{l}{\textbf{Architecture}} \\
Parameters & \multicolumn{2}{c}{82M} \\
Input shape & $25\times1\times128\times128$ & $24\times1\times256\times256$ \\
Encoder stages & \multicolumn{2}{c}{4} \\
Residual blocks / encoder stage & \multicolumn{2}{c}{2} \\
Encoder channels & \multicolumn{2}{c}{[128, 256, 512, 512]} \\
Decoder stages & \multicolumn{2}{c}{4} \\
Residual blocks / decoder stage & \multicolumn{2}{c}{3} \\
Decoder channels & \multicolumn{2}{c}{[512, 512, 256, 128]} \\
Latent channels & \multicolumn{2}{c}{16} \\
Latent resolution & $16\times16$ & $32\times32$ \\
Spatial compression & \multicolumn{2}{c}{$8\times$} \\
Temporal compression & \multicolumn{2}{c}{$1\times$} \\
Normalization & \multicolumn{2}{c}{GroupNorm, 32 groups} \\
Activation & \multicolumn{2}{c}{SiLU} \\
\midrule
\rowcolor{black!10}\multicolumn{3}{l}{\textbf{Training}} \\
\# steps & 1,488,000 & 2,000,000 \\
Optimizer & \multicolumn{2}{c}{AdamW, $(\beta_1,\beta_2)=(0.5,0.9)$} \\
Batch size  & \multicolumn{2}{c}{8}\\
Learning rate & \multicolumn{2}{c}{$10^{-4}$} \\
Weight decay & \multicolumn{2}{c}{0.01} \\
EMA decay & \multicolumn{2}{c}{0.9999} \\
Gradient clipping norm & \multicolumn{2}{c}{1.0} \\
Adversarial warmup steps & \multicolumn{2}{c}{100,000} \\
\bottomrule
\end{tabular}
\end{table}

\subsection{Transformer Architecture, Training, and Inference}
\label{app:transformer_architecture}
Table~\ref{tab:transformer_arch} collects the default architecture, flow-matching pretraining settings, and pretrained-model inference configurations for MRMS and SEVIR. RL post-training settings are reported separately in Appendix~\ref{app:rlvr_impl}.

\begin{table}[!ht]
\appendixtablestyle
\caption{Transformer architecture, flow-matching pretraining, and inference configuration. Parameters include the conditioner, time embedding, and output layer. Inference settings refer to the pretrained-model evaluation configurations.}
\label{tab:transformer_arch}
\begin{tabular}{@{}>{\raggedright\arraybackslash}p{0.36\linewidth}*{2}{>{\centering\arraybackslash}p{0.25\linewidth}}@{}}
\toprule
\textbf{Setting} & \textbf{SEVIR} & \textbf{MRMS} \\
\midrule
\rowcolor{black!10}\multicolumn{3}{l}{\textbf{Architecture}} \\
Parameters & \multicolumn{2}{c}{130M} \\
Frames (history / future) & 5 / 20 & 4 / 20 \\
Transformer blocks & \multicolumn{2}{c}{12} \\
Hidden dimension & \multicolumn{2}{c}{768} \\
Feedforward dimension & \multicolumn{2}{c}{3072} \\
Attention heads & \multicolumn{2}{c}{12} \\
Head dimension & \multicolumn{2}{c}{64} \\
Patch size & \multicolumn{2}{c}{$2\times2$} \\
Tokens per sample & $25\times8\times8=1600$ & $24\times16\times16=6144$ \\
Latent channels & \multicolumn{2}{c}{16} \\
Block normalization & \multicolumn{2}{c}{LayerNorm} \\
QK normalization & \multicolumn{2}{c}{RMSNorm} \\
Positional encoding & \multicolumn{2}{c}{3D RoPE} \\
Time modulation & \multicolumn{2}{c}{adaLN-Zero} \\
FFN activation & \multicolumn{2}{c}{GELU} \\
History fusion & \multicolumn{2}{c}{Additive} \\
\midrule
\rowcolor{black!10}\multicolumn{3}{l}{\textbf{Training}} \\
\# steps & 372,000 & 500,000 \\
Optimizer & \multicolumn{2}{c}{AdamW, $(\beta_1,\beta_2)=(0.9,0.95)$} \\
Batch size  & \multicolumn{2}{c}{32} \\
Learning rate & \multicolumn{2}{c}{$10^{-4}$} \\
Learning rate schedule & \multicolumn{2}{c}{Constant} \\
EMA decay & \multicolumn{2}{c}{0.9999} \\
Gradient clipping norm & \multicolumn{2}{c}{1.0} \\
Time sampler & \multicolumn{2}{c}{Logit-normal, $\mu=-1$, $\sigma=1$} \\
DyPro correlation strength $\alpha$ & \multicolumn{2}{c}{0.5} \\
Condition dropout & \multicolumn{2}{c}{0.1 (zero conditioning)} \\
\midrule
\rowcolor{black!10}\multicolumn{3}{l}{\textbf{Inference}} \\
Sampler & \multicolumn{2}{c}{Euler with DyPro (Algorithm~\ref{alg:code_sample})} \\
Sampling steps & \multicolumn{2}{c}{10} \\
Timestep grid & \multicolumn{2}{c}{Uniform in $[0,1]$} \\
CFG weight (without RL) & \multicolumn{2}{c}{1.6} \\
\bottomrule
\end{tabular}
\end{table}

\paragraph{History Conditioning} Let $\mathbf{z}_{\mathrm{cond}}\in\mathbb{R}^{B\times T\times C\times H\times W}$ denote the conditioning sequence, which places the observed latent frames in the first $L$ positions and masks the remaining future positions with zeros. The conditioner first applies a frame-wise convolutional patch encoder with the same patch size as the noisy-input embedder. It then adds a learnable spatial-topology embedding and a learnable frame-index embedding, followed by a SiLU activation and a zero-initialized $1\times1$ convolution. Denoting the resulting condition features by $\mathbf{c}$ and the noisy patch embeddings by $\mathbf{h}_t$, conditioning is performed through
\begin{equation}
\widetilde{\mathbf{h}}_t = \mathbf{h}_t + \mathbf{c}.
\end{equation}
The conditioned embeddings are then flattened into spatiotemporal tokens and processed by the standard DiT blocks. Thus, the conditioner changes neither the Transformer blocks nor their input--output interface and requires no cross-attention or task-specific fusion pathway.

\paragraph{Classifier-free guidance and Domain Guidance.}
During inference, we adopt classifier-free guidance that adjusts the influence of historical conditioning. 
Classifier-free guidance~\citep{cfg-ho2022classifier} adjusts the influence of historical conditioning by combining conditional and unconditional velocity predictions from the same checkpoint. Let $\theta_{\mathrm{pre}}$ denote the parameters before RL post-training and let $\emptyset$ denote zero conditioning, consistent with the condition-dropout setting in Table~\ref{tab:transformer_arch}. The guided prediction is
\begin{equation}
\mathbf{v}^{\mathrm{CFG}}(\mathbf{z}_t,t\mid\mathbf{c})
=w\,\mathbf{v}_{\theta_{\mathrm{pre}}}(\mathbf{z}_t,t\mid\mathbf{c})
+(1-w)\,\mathbf{v}_{\theta_{\mathrm{pre}}}(\mathbf{z}_t,t\mid\emptyset),
\label{eq:cfg_pretrained}
\end{equation}
where $w$ is the guidance weight and $w=1$ recovers the conditional prediction without guidance. We adopt $w=1.6$ at pretraining stage.



\section{Foundation, Training and Sampling with DyPro}
\label{app:dypro_details}

We describe the noise transform, training target, and deterministic sampling update associated with the DyPro parameterization in Section~\ref{sec:dypro}, followed by their distributional properties and continuous-time interpretation. The stochastic extension used for post-training is presented in Appendix~\ref{app:dypro_sde}. Diffusion time increases from noise at $t=0$ to data at $t=1$. We condition throughout on the history $\mathbf{c}$ and write $\mathbf{y}=\mathbf{z}_1$ for the clean future latent sequence. Each of its $S$ frames has $d$ latent coordinates, so the flattened sequence has dimension $m=Sd$. The Gaussian innovations are independent of $\mathbf{y}$ and $\mathbf{c}$.

\begingroup
\definecolor{codeblue}{rgb}{0.25,0.5,0.5}
\definecolor{codesign}{RGB}{0,0,255}
\definecolor{codefunc}{rgb}{0.85,0.18,0.50}
\definecolor{highlightcolor}{HTML}{F2F2F2}
\lstdefinelanguage{AppendixPython}{
  language=Python,
  morekeywords=[2]{repeat_group,no_grad,shared_noise,
    dypro_sde_step,sde_step,ode_step,decode,denormalize,R_struct,R_detail,
    R_noise,R_clean,
    group_normalize,clipped_mixgrpo,grpo_update,update},
  keywordstyle=\color{blue}\bfseries,
  keywordstyle=[2]\color{codefunc},
  commentstyle=\color{codeblue},
  stringstyle=\color{orange},
  literate=
    {*}{{\color{codesign}*}}{1}
    {-}{{\color{codesign}-}}{1}
    {+}{{\color{codesign}+}}{1}
    {/}{{\color{codesign}/}}{1},
}
\lstset{
  language=AppendixPython,
  backgroundcolor=\color{white},
  basicstyle=\fontsize{8pt}{8.4pt}\ttfamily\selectfont,
  columns=fullflexible,
  keepspaces=true,
  showstringspaces=false,
  breaklines=true,
  captionpos=b,
  aboveskip=4pt,
  belowskip=4pt,
}

\lstset{deletekeywords={range},moreemph={range,copy}}
\begin{algorithm}[t]
    \caption{DyPro Noise Correlation and Decorrelation}
    \label{alg:dypro_correlation}
\begin{lstlisting}[linebackgroundcolor={%
    \ifnum\value{lstnumber}=11 \color{highlightcolor}\fi
    \ifnum\value{lstnumber}=21 \color{highlightcolor}\fi
}]
# xi, e: frame sequences; s indexes frames
# t: diffusion timestep; alpha: shared DyPro strength

def correlate(xi, t):
    a = alpha * (1 - t)
    q = 1 / sqrt(1 + a**2)
    r = a * q

    e = xi.copy()
    for s in range(1, len(xi)):
        e[s] = r * e[s - 1] + q * xi[s]
    return e

def decorrelate(e, t):
    a = alpha * (1 - t)
    q = 1 / sqrt(1 + a**2)
    r = a * q

    xi = e.copy()
    for s in range(1, len(e)):
        xi[s] = (e[s] - r * e[s - 1]) / q
    return xi
\end{lstlisting}
\end{algorithm}
\endgroup

\subsection{Correlated Noise}
\label{app:dypro_noise_transform}
For finite $\alpha\geq0$, define
\begin{equation}
r_t=\sqrt{\gamma_t}=\frac{\alpha(1-t)}{\sqrt{1+\alpha^2(1-t)^2}},
\qquad
q_t=\sqrt{1-\gamma_t}=\frac{1}{\sqrt{1+\alpha^2(1-t)^2}}.
\label{eq:dypro_rq}
\end{equation}

The frame-wise recursion defined in \Eqref{eq:dypro_noise_def}, practically implemented through \texttt{correlate} in Algorithm~\ref{alg:dypro_correlation}, denotes a lower-triangular matrix representation. With frame indices $s,j\in\{0,\ldots,S-1\}$, let
\begin{equation}
(C_t)_{sj}=\begin{cases}
r_t^s, & j=0,\\
q_t r_t^{s-j}, & 1\leq j\leq s,\\
0, & j>s.
\end{cases}
\qquad
\mathcal{C}_t=C_t\otimes\mathbf{I}_d,
\qquad
\boldsymbol{\epsilon}_t=\mathcal{C}_t\boldsymbol{\xi},
\label{eq:dypro_transform}
\end{equation}
where $\boldsymbol{\xi}\sim\mathcal{N}(\mathbf{0},\mathbf{I}_m)$ and $r_t^0=1$, including when $r_t=0$.

Because $q_t>0$, this transform is invertible. Its inverse $\mathcal{C}_t^{-1}=C^{-1}\otimes\mathbf{I}_d$, implemented by \texttt{decorrelate} in Algorithm~\ref{alg:dypro_correlation}, follows directly from the recursion:
\begin{equation}
\boldsymbol{\xi}^1=\boldsymbol{\epsilon}_t^1,
\qquad
\boldsymbol{\xi}^s=\frac{\boldsymbol{\epsilon}_t^s-r_t\boldsymbol{\epsilon}_t^{s-1}}{q_t},
\quad s\geq2.
\label{eq:dypro_inverse}
\end{equation}

The joint noise $\boldsymbol{\epsilon}_t$ is Gaussian because it is a linear transform of independent Gaussian innovations. Since $r_t^2+q_t^2=1$, the recursion preserves the standard Gaussian marginal of each frame. Independence of subsequent innovations gives $\operatorname{Cov}(\boldsymbol{\epsilon}_t^s,\boldsymbol{\epsilon}_t^j)=r_t^{|s-j|}\mathbf{I}_d=\gamma_t^{|s-j|/2}\mathbf{I}_d$, yielding the full covariance and distribution
\begin{equation}
\mathbf{Q}_t=\mathcal{C}_t\mathcal{C}_t^{\top}
=\big[r_t^{|s-j|}\big]_{s,j=0}^{S-1}\otimes\mathbf{I}_d,
\qquad
\boldsymbol{\epsilon}_t\sim\mathcal{N}(\mathbf{0},\mathbf{Q}_t).
\label{eq:dypro_joint_noise}
\end{equation}
For $\alpha=0$, or at $t=1$, $r_t=0$ and $q_t=1$, so $\mathcal{C}_t=\mathbf{I}_m$ and the noise is independent across frames.

\subsection{Linear Interpolation under Correlated Noise}
\label{app:dypro_properties}
The linear interpolation between $\mathbf{y}$ and
$\boldsymbol{\epsilon}_t$ incorporates the time dependence of the
noise transform. Holding $\mathbf{y}$ and $\boldsymbol{\xi}$ fixed,
the interpolation path and its derivative are
\begin{align}
\mathbf{z}_t
&= t\mathbf{y}+(1-t)\boldsymbol{\epsilon}_t
= t\mathbf{y}+(1-t)\mathcal{C}_t\boldsymbol{\xi},
\label{eq:dypro_path}\\
\dot{\mathbf{z}}_t
&= \mathbf{y}-\boldsymbol{\epsilon}_t
+(1-t)\dot{\mathcal{C}}_t\boldsymbol{\xi}.
\label{eq:dypro_path_derivative}
\end{align}
Consequently, for $t<1$, the conditional distribution along this path is
\begin{equation}
p_t(\mathbf{z}\mid\mathbf{y},\mathbf{c})
=
\mathcal{N}\!\left(
\mathbf{z};
t\mathbf{y},
(1-t)^2\mathbf{Q}_t
\right).
\label{eq:dypro_conditional_path}
\end{equation}
Marginalizing over $\mathbf{y}\sim p_{\mathrm{data}}(\cdot\mid\mathbf{c})$
defines a probability path from
$p_0=\mathcal{N}(\mathbf{0},\mathbf{Q}_0)$ to
$p_1=p_{\mathrm{data}}(\cdot\mid\mathbf{c})$.

To express the path derivative in terms of the correlated noise, define
\begin{equation}
\mathbf{B}_t
=
\dot{\mathcal{C}}_t\mathcal{C}_t^{-1},
\qquad
\dot{\boldsymbol{\epsilon}}_t
=
\mathbf{B}_t\boldsymbol{\epsilon}_t.
\label{eq:dypro_noise_velocity}
\end{equation}
The operator $\mathbf{B}_t$ describes how the noise transform changes
with diffusion time for fixed innovations. Averaging
\Eqref{eq:dypro_path_derivative} over the conditional distribution
of the clean sequence and innovations gives the marginal velocity
\begin{align}
\mathbf{v}_t(\mathbf{z}\mid\mathbf{c})
&=
\mathbb{E}\!\left[
\dot{\mathbf{z}}_t
\mid \mathbf{z}_t=\mathbf{z},\mathbf{c}
\right]
\nonumber\\
&=
\mathbb{E}\!\left[
\mathbf{y}-\boldsymbol{\epsilon}_t
\mid \mathbf{z}_t=\mathbf{z},\mathbf{c}
\right]
+
(1-t)\mathbf{B}_t
\mathbb{E}\!\left[
\boldsymbol{\epsilon}_t
\mid \mathbf{z}_t=\mathbf{z},\mathbf{c}
\right].
\label{eq:dypro_marginal_velocity}
\end{align}
This velocity satisfies the continuity equation
$\partial_t p_t+\nabla_{\mathbf{z}}\cdot(p_t\mathbf{v}_t)=0$.
The associated deterministic generative dynamics are therefore
\begin{equation}
\frac{\mathrm{d}\mathbf{z}_t}{\mathrm{d}t}
=
\mathbf{v}_t(\mathbf{z}_t\mid\mathbf{c}),
\qquad
\mathbf{z}_0=\mathcal{C}_0\boldsymbol{\xi},
\quad
\boldsymbol{\xi}\sim\mathcal{N}(\mathbf{0},\mathbf{I}_m).
\label{eq:dypro_population_ode}
\end{equation}

\subsection{Training Target and Deterministic Sampling}
\label{app:dypro_deterministic}
Algorithm~\ref{alg:code_train} regresses
$\mathbf{y}-\boldsymbol{\epsilon}_t$, providing a simple
parameterization of the marginal velocity in
\Eqref{eq:dypro_marginal_velocity}.
Given one network prediction
$\mathbf{f}_\theta=\mathbf{f}_\theta(\mathbf{z}_t,t\mid\mathbf{c})$,
the clean latent, correlated noise, and innovations are estimated as
\begin{equation}
\begin{aligned}
\widehat{\mathbf{y}}_\theta
&= \mathbf{z}_t+(1-t)\mathbf{f}_\theta,\\
\widehat{\boldsymbol{\epsilon}}_\theta
&= \mathbf{z}_t-t\mathbf{f}_\theta,\\
\widehat{\boldsymbol{\xi}}_\theta
&= \mathcal{C}_t^{-1}\widehat{\boldsymbol{\epsilon}}_\theta.
\end{aligned}
\label{eq:dypro_predictions}
\end{equation}
These identities follow directly from the interpolation path:
$\mathbf{y}=\mathbf{z}_t+(1-t)(\mathbf{y}-\boldsymbol{\epsilon}_t)$
and
$\boldsymbol{\epsilon}_t=\mathbf{z}_t-t(\mathbf{y}-\boldsymbol{\epsilon}_t)$.
Substituting these estimates into
\Eqref{eq:dypro_marginal_velocity} gives the velocity estimate
\begin{equation}
\widehat{\mathbf{v}}_\theta(\mathbf{z}_t,t\mid\mathbf{c})
=
\mathbf{f}_\theta
+
(1-t)\mathbf{B}_t\widehat{\boldsymbol{\epsilon}}_\theta
=
\mathbf{f}_\theta
+
(1-t)\dot{\mathcal{C}}_t
\widehat{\boldsymbol{\xi}}_\theta.
\label{eq:dypro_learned_velocity}
\end{equation}

For deterministic sampling, we integrate this velocity by holding
the clean estimate $\widehat{\mathbf{y}}_\theta$ fixed within each
step from $t$ to $t'>t$. Using
\Eqref{eq:dypro_predictions}, the velocity can be written as
\begin{equation}
\widehat{\mathbf{v}}_\theta(\mathbf{z},t\mid\mathbf{c})
=
\frac{\widehat{\mathbf{y}}_\theta-\mathbf{z}}{1-t}
+
\mathbf{B}_t
\left(\mathbf{z}-t\widehat{\mathbf{y}}_\theta\right).
\label{eq:dypro_clean_parameterized_velocity}
\end{equation}
For $\tau\in[t,t')$, define the residual
$\mathbf{w}_\tau=\mathbf{z}_\tau-\tau\widehat{\mathbf{y}}_\theta$.
With the clean estimate fixed, its dynamics satisfy
\begin{equation}
\frac{\mathrm{d}\mathbf{w}_\tau}{\mathrm{d}\tau}
=
\left(
\mathbf{B}_\tau-\frac{\mathbf{I}_m}{1-\tau}
\right)\mathbf{w}_\tau.
\label{eq:dypro_residual_ode}
\end{equation}
Since
\begin{equation}
\frac{\mathrm{d}}{\mathrm{d}\tau}
\left[(1-\tau)\mathcal{C}_\tau\right]
=
\left(
\mathbf{B}_\tau-\frac{\mathbf{I}_m}{1-\tau}
\right)
(1-\tau)\mathcal{C}_\tau,
\end{equation}
the residual has the closed-form update
\begin{equation}
\mathbf{w}_{t'}
=
\frac{1-t'}{1-t}
\mathcal{C}_{t'}\mathcal{C}_t^{-1}\mathbf{w}_t.
\label{eq:dypro_residual_update}
\end{equation}
Substituting
$\mathbf{w}_t=(1-t)\widehat{\boldsymbol{\epsilon}}_\theta$
yields the deterministic sampling rule
\begin{align}
\mathbf{z}_{t'}
&=
t'\widehat{\mathbf{y}}_\theta
+
(1-t')\mathcal{C}_{t'}\mathcal{C}_t^{-1}
\widehat{\boldsymbol{\epsilon}}_\theta
\nonumber\\
&=
t'\widehat{\mathbf{y}}_\theta
+
(1-t')\mathcal{C}_{t'}\widehat{\boldsymbol{\xi}}_\theta.
\label{eq:dypro_deterministic_update}
\end{align}
The update is exact for the within-step dynamics with a fixed clean
estimate and is first-order consistent with the learned ODE.
It extends continuously to $t'=1$, where it returns
$\widehat{\mathbf{y}}_\theta$.

Algorithm~\ref{alg:code_sample} implements this update by applying
\texttt{decorrelate} at $t$ and \texttt{correlate} at $t'$.
Each step uses one evaluation of $\mathbf{f}_\theta$ and transports
the estimated innovations to the next timestep. For $\alpha=0$,
$\mathcal{C}_t=\mathbf{I}_m$ throughout, and the update reduces to
$\mathbf{z}_{t'}=\mathbf{z}_t+(t'-t)\mathbf{f}_\theta$.

\subsection{Stochastic Sampling}
\label{app:dypro_sde}

\begingroup
\definecolor{codeblue}{rgb}{0.25,0.5,0.5}
\definecolor{codesign}{RGB}{0,0,255}
\definecolor{codefunc}{rgb}{0.85,0.18,0.50}
\definecolor{highlightcolor}{HTML}{F2F2F2}
\lstdefinelanguage{AppendixPython}{
  language=Python,
  morekeywords=[2]{repeat_group,no_grad,shared_noise,
    dypro_sde_step,sde_step,ode_step,decode,denormalize,R_struct,R_detail,
    R_noise,R_clean,
    group_normalize,clipped_mixgrpo,grpo_update,update},
  keywordstyle=\color{blue}\bfseries,
  keywordstyle=[2]\color{codefunc},
  commentstyle=\color{codeblue},
  stringstyle=\color{orange},
  literate=
    {*}{{\color{codesign}*}}{1}
    {-}{{\color{codesign}-}}{1}
    {+}{{\color{codesign}+}}{1}
    {/}{{\color{codesign}/}}{1},
}
\lstset{
  language=AppendixPython,
  backgroundcolor=\color{white},
  basicstyle=\fontsize{8pt}{8.4pt}\ttfamily\selectfont,
  columns=fullflexible,
  keepspaces=true,
  showstringspaces=false,
  breaklines=true,
  captionpos=b,
  aboveskip=4pt,
  belowskip=4pt,
}

\lstset{morekeywords=[2]{net,decorrelate,correlate,sqrt,randn_like,stopgrad,
    sum_latent_dims,normal_log_prob}}
\begin{algorithm}[t]
    \caption{DyPro SDE Sampling (w/o CFG)}
    \label{alg:dypro_sde}
\begin{lstlisting}[linebackgroundcolor={%
    \ifnum\value{lstnumber}=7 \color{highlightcolor}\fi
    \ifnum\value{lstnumber}>9 \ifnum\value{lstnumber}<12 \color{highlightcolor}\fi\fi
    \ifnum\value{lstnumber}>14 \ifnum\value{lstnumber}<18 \color{highlightcolor}\fi\fi
}]
# One active SDE transition: t -> t_next < 1
# kappa: stochasticity; 0 < kappa**2 * (t_next - t) <= 1

v = net(z, t, c)
x = z + (1 - t) * v
e = z - t * v
e_init = decorrelate(e, t)

a = kappa * sqrt(t_next - t)
e_mean = sqrt(1 - a**2) * e_init
e = correlate(e_mean + a * randn_like(e_init), t_next)
z_next = t_next * x + (1 - t_next) * e

mu = t_next * x + (1 - t_next) * correlate(e_mean, t_next)
residual = decorrelate(stopgrad(z_next) - mu, t_next)
scale = (1 - t_next) * a
log_score = avg_latent_dims(normal_log_prob(residual, 0, scale))

return z_next, log_score
\end{lstlisting}
\end{algorithm}
\endgroup

A stochastic extension of the deterministic dynamics refreshes the
noise in independent-innovation coordinates. Let $\kappa_t\geq0$
be a prescribed stochasticity schedule independent of the policy
parameters. We replace the fixed innovations with the
Ornstein--Uhlenbeck process
\begin{equation}
\mathrm{d}\boldsymbol{\xi}_t
=
-\frac{\kappa_t^2}{2}\boldsymbol{\xi}_t\,\mathrm{d}t
+
\kappa_t\,\mathrm{d}\mathbf{W}_t,
\qquad
\boldsymbol{\xi}_0\sim
\mathcal{N}(\mathbf{0},\mathbf{I}_m),
\label{eq:dypro_innovation_sde}
\end{equation}
where $\mathbf{W}_t$ is an $m$-dimensional standard Brownian motion,
independent of the clean sequence and history.
This process preserves the standard Gaussian distribution of the
innovations at every timestep. Consequently,
$\mathbf{z}_t=t\mathbf{y}+(1-t)\mathcal{C}_t\boldsymbol{\xi}_t$
retains the conditional distributions in
\Eqref{eq:dypro_conditional_path}.

Applying the product rule, taking conditional expectations given
$(\mathbf{z}_t,\mathbf{c})$, and parameterizing them with the network
predictions gives
\begin{equation}
\mathrm{d}\mathbf{z}_t
=
\left[
\mathbf{f}_\theta
+
(1-t)\left(
\mathbf{B}_t-\frac{\kappa_t^2}{2}\mathbf{I}_m
\right)
\widehat{\boldsymbol{\epsilon}}_\theta
\right]\mathrm{d}t
+
\kappa_t(1-t)\mathcal{C}_t\,\mathrm{d}\mathbf{W}_t.
\label{eq:dypro_sampling_sde}
\end{equation}
Here, the network prediction and noise estimate are evaluated at
$(\mathbf{z}_t,t,\mathbf{c})$.
Setting $\kappa_t=0$ recovers
the deterministic dynamics.

For a sampling step from $t$ to $t'>t$ with $t'<1$, we hold
the clean estimate $\widehat{\mathbf{y}}_\theta$ fixed and discretize
the innovation process in \Eqref{eq:dypro_innovation_sde} from
$\widehat{\boldsymbol{\xi}}_\theta$ using $h=t'-t$ and
\begin{equation}
a=\kappa_t\sqrt{h},
\qquad
b=\sqrt{1-a^2},
\qquad
0\leq a\leq1.
\label{eq:dypro_stochastic_coefficients}
\end{equation}
Here, $a^2+b^2=1$ preserves the standard Gaussian reference
distribution in innovation coordinates. For an independent
$\boldsymbol{\eta}\sim\mathcal{N}(\mathbf{0},\mathbf{I}_m)$,
the resulting update is
\begin{equation}
\mathbf{z}_{t'}
=
t'\widehat{\mathbf{y}}_\theta
+
(1-t')\mathcal{C}_{t'}
\left(
b\mathcal{C}_t^{-1}
\widehat{\boldsymbol{\epsilon}}_\theta
+
a\boldsymbol{\eta}
\right).
\label{eq:dypro_stochastic_update}
\end{equation}
All network estimates are evaluated at the current state
$(\mathbf{z}_t,t,\mathbf{c})$.
Algorithm~\ref{alg:dypro_sde} implements this update by
decorrelating the predicted noise, mixing the resulting innovations
with independent Gaussian noise, and applying the noise transform
at $t'$.
For $a=0$, the update reduces to
\Eqref{eq:dypro_deterministic_update}.

\section{Timestep-aware post-training}
\label{app:rlvr}

Post-training combines a terminal forecast reward with policy updates on selected stochastic transitions. We describe mixed rollouts, timestep-aware reward evaluation, and the local policy objective, together with the training algorithm and configuration.

\subsection{Mixed Sampling and Window Scheduling}
\label{app:mixed_sampling}
We discretize diffusion time as $0=t_0<\cdots<t_N=1$ and activate a contiguous window $\mathcal{W}_{\ell}=\{\ell,\ldots,\ell+K-1\}$, with default size $K=4$. The rollout policy $\theta_{\mathrm{old}}$ is held fixed while collecting the group. Transitions inside the window use the stochastic kernel from Appendix~\ref{app:dypro_sde}, while the remaining transitions use the deterministic update from Appendix~\ref{app:dypro_deterministic}:
\begin{equation}
\mathbf{z}_{t_{n+1}} =
\begin{cases}
\operatorname{Sample}\!\left[p^{\mathrm{DyPro}}_{\theta_{\mathrm{old}}}
(\,\cdot\mid\mathbf{z}_{t_n},\mathbf{c})\right],
& n\in\mathcal{W}_{\ell}, \\
\Phi^{\mathrm{DyPro}}_{\theta_{\mathrm{old}}}(\mathbf{z}_{t_n},t_n,t_{n+1},\mathbf{c}),
& n\notin\mathcal{W}_{\ell}.
\end{cases}
\label{eq:mixed_ode_sde}
\end{equation}
Here, $p^{\mathrm{DyPro}}$ is the transition distribution induced by Equation~\ref{eq:dypro_stochastic_update}, and $\Phi^{\mathrm{DyPro}}$ denotes the deterministic update in Equation~\ref{eq:dypro_deterministic_update}. The trajectories share an initial DyPro latent and conditioning context, with independent stochastic innovations inside the window.

The stochastic window contains only nonterminal transitions, with
$0<\kappa_{t_n}^2(t_{n+1}-t_n)\leq1$ for each active step. Its bounds are
\begin{equation}
1\leq K\leq N-1,\qquad
0\leq\ell\leq\ell_{\max}=N-K-1.
\label{eq:dypro_window_bound}
\end{equation}
The final transition to $t_N=1$ is deterministic. We start at $\ell=0$
and cycle through window starts spaced by $s_w$, switching after every
$M_{\mathrm{shift}}$ training iterations:
\begin{equation}
\ell\leftarrow
\begin{cases}
\ell+s_w, & \ell+s_w\leq\ell_{\max},\\
0, & \text{otherwise}.
\end{cases}
\label{eq:window_schedule}
\end{equation}
With $N=10$, $K=4$, $s_w=2$, and $M_{\mathrm{shift}}=20$ throughout post-training.

\subsection{Timestep-aware reward evaluation}
\label{app:reward_details}
Each rollout produces a final decoded forecast $\hat{\mathbf{x}}^i=D(\mathbf{z}_1^i)$, which is compared with the reference future sequence $\mathbf{x}^*$. The two reward components are defined in Equations~\ref{eq:noise_reward} and~\ref{eq:clean_reward}: $R_{\mathrm{noise}}$ combines pooled low-threshold CSI with SSIM, and $R_{\mathrm{clean}}$ combines high-threshold CSI with negative LPIPS. Local max pooling applies only to the low-threshold CSI terms; the other metrics are evaluated on the decoded forecast without this pooling operation. Metric definitions are given in Appendix~\ref{app:evaluation_metrics}.

For an active transition $n$, the terminal reward assigned to trajectory $i$ is $R_n^i=R(\hat{\mathbf{x}}^i,\mathbf{x}^*;t_n)$, as defined in Equation~\ref{eq:stage_aware_reward}. All active transitions are evaluated on the same final forecast. Their metric weights depend on their own diffusion times $t_n$, so rewards and group-relative advantages are computed separately for each transition. No reward is evaluated on an intermediate noisy latent.

\subsection{MixGRPO Objective and Policy Scores}
\label{app:grpo_math}
For each active transition $n$, the group provides a relative measure of forecast quality. We normalize its terminal rewards across the $G$ trajectories:
\begin{equation}
\hat{A}^i_n =
\frac{R_n^i-\operatorname{mean}(\{R_n^j\}_{j=1}^{G})}
{\operatorname{std}(\{R_n^j\}_{j=1}^{G})+\varepsilon_A},
\label{eq:group_advantage}
\end{equation}
where $\varepsilon_A>0$ stabilizes the denominator. The comparison is made across forecasts at the same transition, rather than across diffusion times. If all group rewards agree, their advantages are zero.

For a stochastic step from $t$ to $t'<1$ with $a>0$, the update in
Equation~\ref{eq:dypro_stochastic_update} defines the conditional density
\begin{align}
p^{\mathrm{DyPro}}_\theta(\mathbf{z}'\mid\mathbf{z},\mathbf{c})
&=\mathcal{N}\big(\mathbf{z}';\boldsymbol{\mu}_\theta,
\sigma^2\mathbf{Q}_{t'}\big),\\
\boldsymbol{\mu}_\theta
&=t'\widehat{\mathbf{y}}_\theta
+(1-t')b\mathcal{C}_{t'}\mathcal{C}_t^{-1}
\widehat{\boldsymbol{\epsilon}}_\theta,
\qquad \sigma=(1-t')a.
\label{eq:dypro_transition_parameters}
\end{align}
All network estimates are evaluated at $(\mathbf{z},t,\mathbf{c})$;
the transition times are implicit in $p^{\mathrm{DyPro}}_\theta$.
Whitening the residual as
$\mathbf{w}_\theta=\mathcal{C}_{t'}^{-1}
(\mathbf{z}'-\boldsymbol{\mu}_\theta)$ gives the joint log density
\begin{equation}
\log p^{\mathrm{DyPro}}_\theta(\mathbf{z}'\mid\mathbf{z},\mathbf{c})
=-\frac{m}{2}\log(2\pi)-m\log\sigma
-\log|\det\mathcal{C}_{t'}|
-\frac{\|\mathbf{w}_\theta\|_2^2}{2\sigma^2}.
\label{eq:dypro_logdensity}
\end{equation}
For trajectory $i$ at transition $n$, we write
$\log p_{\theta,n}^i
=\log p^{\mathrm{DyPro}}_\theta
(\mathbf{z}_{t_{n+1}}^i\mid\mathbf{z}_{t_n}^i,\mathbf{c})$,
with $\log p_{\mathrm{old},n}^i$ defined analogously for
$\theta_{\mathrm{old}}$.

For a stored state and successor from the rollout, the policy ratio is
\begin{equation}
\rho^i_n(\theta)=
\frac{p^{\mathrm{DyPro}}_{\theta}(\mathbf{z}^i_{t_{n+1}}\mid\mathbf{z}^i_{t_n},\mathbf{c})}
{p^{\mathrm{DyPro}}_{\theta_{\mathrm{old}}}(\mathbf{z}^i_{t_{n+1}}\mid\mathbf{z}^i_{t_n},\mathbf{c})}
=\exp\big(\log p^i_{\theta,n}-\log p^i_{\mathrm{old},n}\big),
\label{eq:rlvr_ratio}
\end{equation}
Both policies are evaluated on the same stored transition. The states, terminal rewards, advantages, and old-policy scores are held fixed during the update; gradients flow through the current-policy log density.

When both policies share the time grid, $\alpha$, and stochasticity schedule, their covariance and whitening Jacobian are identical and independent of $\theta$. Let $\mathbf{w}^i_{\theta,n}$ and $\sigma_n$ denote the residual and scale in Equation~\ref{eq:dypro_logdensity} for trajectory $i$ at transition $n$. Cancelling the shared terms in the two log densities yields
\begin{equation}
\log\rho^i_n(\theta)
=\frac{\|\mathbf{w}^i_{\mathrm{old},n}\|_2^2-\|\mathbf{w}^i_{\theta,n}\|_2^2}{2\sigma_n^2}.
\label{eq:rl_logratio_residual}
\end{equation}
The Jacobian can thus be omitted from a score used only in this difference, although it remains part of the absolute joint log density.

Following MixGRPO~\citep{li2025mixgrpo} and Flow-GRPO~\citep{liu2025flowgrpo}, we maximize the clipped local surrogate
\begin{multline}
\mathcal{J}_{\mathrm{MixGRPO}}(\theta)=
\mathbb{E}\Bigg[
\frac{1}{G|\mathcal{W}_{\ell}|}\sum_{i=1}^{G}
\sum_{n\in\mathcal{W}_{\ell}}
\min\Big(
\rho^i_n(\theta)\hat{A}^i_n,
\operatorname{clip}(\rho^i_n(\theta),1-\varepsilon_c,1+\varepsilon_c)
\hat{A}^i_n\Big)
\Bigg],
\label{eq:mixgrpo_objective}
\end{multline}
where the expectation is over conditioning examples and old-policy rollouts, $|\mathcal{W}_{\ell}|=K$, and $\varepsilon_c$ is the clipping range. The implemented minimization loss is the negative of this objective. Only the stochastic transitions participate in the surrogate; deterministic transitions still influence the sampled final forecast.

\subsection{Training Algorithm}
\label{app:rl_algorithms}
Algorithm~\ref{alg:stage_aware_rlvr} presents the full training loop. The helper \texttt{clipped\_mixgrpo} optimizes the negative of Equation~\ref{eq:mixgrpo_objective}, holding the rollout data fixed. Old-policy scores are cached internally, following the convention in Appendix~\ref{app:grpo_math}.

The stochastic helper \texttt{dypro\_sde\_step} implements Equation~\ref{eq:dypro_stochastic_update} and obtains the next time from the grid. The deterministic helper \texttt{ode\_step} denotes $\Phi^{\mathrm{DyPro}}$, and \texttt{max\_left} is the bound in Equation~\ref{eq:dypro_window_bound}.

The helper \texttt{shared\_noise(G)} repeats one DyPro initial latent across the group. Both reward helpers include the specified metric coefficients, and \texttt{group\_normalize} includes $\varepsilon_A$. Contexts and targets are broadcast as needed. Latent denormalization is included in \texttt{decode}.

\begingroup
\definecolor{codeblue}{rgb}{0.25,0.5,0.5}
\definecolor{codesign}{RGB}{0,0,255}
\definecolor{codefunc}{rgb}{0.85,0.18,0.50}
\definecolor{highlightcolor}{HTML}{F2F2F2}
\lstdefinelanguage{AppendixPython}{
  language=Python,
  morekeywords=[2]{repeat_group,no_grad,shared_noise,
    dypro_sde_step,sde_step,ode_step,decode,denormalize,R_struct,R_detail,
    R_noise,R_clean,
    group_normalize,clipped_mixgrpo,grpo_update,update},
  keywordstyle=\color{blue}\bfseries,
  keywordstyle=[2]\color{codefunc},
  commentstyle=\color{codeblue},
  stringstyle=\color{orange},
  literate=
    {*}{{\color{codesign}*}}{1}
    {-}{{\color{codesign}-}}{1}
    {+}{{\color{codesign}+}}{1}
    {/}{{\color{codesign}/}}{1},
}
\lstset{
  language=AppendixPython,
  backgroundcolor=\color{white},
  basicstyle=\fontsize{8pt}{8.4pt}\ttfamily\selectfont,
  columns=fullflexible,
  keepspaces=true,
  showstringspaces=false,
  breaklines=true,
  captionpos=b,
  aboveskip=4pt,
  belowskip=4pt,
}

\begin{algorithm}[H]
\caption{Stage-Aware Post-Training with Window Scheduling}
\label{alg:stage_aware_rlvr}
\begin{lstlisting}
# theta: trainable policy initialized from the flow-matching checkpoint
# K: active-window size (default: 4); max_left: last valid window start

left = 0
for iteration, (c, x_gt) in enumerate(dataloader):
    window = range(left, left + K)
    traj[:, 0] = shared_noise(G)

    for n, t in enumerate(timesteps[:-1]):
        if n in window:
            traj[:, n + 1] = dypro_sde_step(theta, traj[:, n], t, c)
        else:
            traj[:, n + 1] = ode_step(theta, traj[:, n], t, c)

    x_sample = decode(traj[:, -1])
    r_noise = R_noise(x_sample, x_gt)
    r_clean = R_clean(x_sample, x_gt)

    for n in window:
        t = timesteps[n]
        rewards[:, n] = (1 - t) * r_noise + t * r_clean

    advantages = group_normalize(rewards[:, window], dim=0)
    loss = clipped_mixgrpo(theta, traj, window, advantages)
    update(theta, loss)

    if (iteration + 1) % shift_interval == 0:
        left += window_stride
        if left > max_left:
            left = 0
\end{lstlisting}
\end{algorithm}
\endgroup

\subsection{Domain Guidance}
\label{app:domain_guidance}
After skill-aware RL post-training, we adopt Domain Guidance (DoG)~\citep{zhong2025domain}, which combines an adapted conditional model with the original pretrained model as the unconditional reference. In our setting, the conditional branch uses the RL-adapted parameters $\theta_{\mathrm{RL}}$ with history $\mathbf{c}$, while the unconditional branch uses the frozen pre-RL parameters $\theta_{\mathrm{pre}}$ with zero conditioning:
\begin{equation}
\mathbf{v}^{\mathrm{DoG}}(\mathbf{z}_t,t\mid\mathbf{c})
=w\,\mathbf{v}_{\theta_{\mathrm{RL}}}(\mathbf{z}_t,t\mid\mathbf{c})
+(1-w)\,\mathbf{v}_{\theta_{\mathrm{pre}}}(\mathbf{z}_t,t\mid\emptyset).
\label{eq:domain_guidance}
\end{equation}
This retains the pretrained unconditional reference while incorporating skill alignment through the conditional branch. Setting $w=1$ recovers the RL model without guidance; using $\theta_{\mathrm{pre}}$ for both branches recovers Equation~\ref{eq:cfg_pretrained}. Post-RL evaluation settings use $w=1.4$, as listed in Tables~\ref{tab:training_configuration}. Guidance is disabled during RL rollouts and applied only at inference. RL+CFG configuration in Appendix~\ref{app:component_ablation} uses this DoG formulation.

\subsection{Post-Training and Reward Configuration}
\label{app:rlvr_impl}

Table~\ref{tab:training_configuration} summarizes the post-training,
reward, and evaluation configurations. The pretrained VAE remains
frozen throughout post-training. Each rollout batch is used for one
gradient update, followed by synchronization of the rollout policy.

The CSI coefficients incorporate uniform averaging within each
threshold subset and the mixing weights for SSIM or negative LPIPS.
No additional normalization is applied to the weighted sums in
Equations~\ref{eq:noise_reward} and~\ref{eq:clean_reward}.
The listed checkpoints identify the models used for evaluation,
rather than the total training budgets.

\begin{table}[!ht]
\appendixtablestyle
\caption{Post-training, reward, and evaluation configurations.
Reward thresholds are expressed in VIL values for SEVIR and
$\mathrm{mm\,h}^{-1}$ for MRMS.}
\label{tab:training_configuration}
\begin{tabular}{@{}>{\raggedright\arraybackslash}p{0.36\linewidth}*{2}{>{\centering\arraybackslash}p{0.25\linewidth}}@{}}
\toprule
\textbf{Setting} & \textbf{SEVIR} & \textbf{MRMS} \\
\midrule
\rowcolor{black!10}\multicolumn{3}{l}{\textbf{Optimization}} \\
Optimizer
    & \multicolumn{2}{c}{AdamW, $(\beta_1,\beta_2)=(0.9,0.95)$} \\
Learning rate
    & \multicolumn{2}{c}{$10^{-5}$, constant} \\
Batch size
    & \multicolumn{2}{c}{8} \\
Policy ratio clipping range $\varepsilon_c$
    & \multicolumn{2}{c}{$10^{-4}$} \\
Advantage clipping
    & \multicolumn{2}{c}{$[-5,5]$} \\
\midrule
\rowcolor{black!10}\multicolumn{3}{l}{\textbf{Rollout}} \\
Group size $G$
    & \multicolumn{2}{c}{16} \\
Sampling steps $N$
    & \multicolumn{2}{c}{10} \\
Stochastic window size $K$
    & \multicolumn{2}{c}{4} \\
Window stride $s_w$
    & \multicolumn{2}{c}{2} \\
Window shift interval $M_{\mathrm{shift}}$
    & \multicolumn{2}{c}{20 training iterations (cyclic)} \\
DyPro correlation strength $\alpha$
    & \multicolumn{2}{c}{0.5} \\
CFG during rollout
    & \multicolumn{2}{c}{Disabled} \\
\midrule
\rowcolor{black!10}\multicolumn{3}{l}{\textbf{Reward}} \\
Low-threshold set $\mathcal{T}_{\mathrm{low}}$
    & $\{16,74,133\}$ & $\{1,2,4\}$ \\
High-threshold set $\mathcal{T}_{\mathrm{high}}$
    & $\{160,181,219\}$ & $\{8,16,32\}$ \\
Low-threshold CSI max pooling
    & \multicolumn{2}{c}{$4\times4$} \\
SSIM coefficient $\lambda_s$
    & \multicolumn{2}{c}{0.5} \\
LPIPS coefficient $\lambda_p$
    & \multicolumn{2}{c}{0.9} \\
CSI coefficient $\lambda_\tau$, $\tau\in\mathcal{T}_{\mathrm{low}}$
    & \multicolumn{2}{c}{$(1-\lambda_s)/|\mathcal{T}_{\mathrm{low}}|$} \\
CSI coefficient $\lambda_\tau$, $\tau\in\mathcal{T}_{\mathrm{high}}$
    & \multicolumn{2}{c}{$(1-\lambda_p)/|\mathcal{T}_{\mathrm{high}}|$} \\
\midrule
\rowcolor{black!10}\multicolumn{3}{l}{\textbf{Evaluation}} \\
\# steps     & 1,000 & 4,000 \\
Sampling
    & \multicolumn{2}{c}{10 Euler steps} \\
CFG weight
    & \multicolumn{2}{c}{1.4} \\
\bottomrule
\end{tabular}
\end{table}

\section{Datasets and Evaluation Protocol}
\label{app:experimental_details}

\subsection{Datasets and Preprocessing}
\label{app:data_details}
Table~\ref{tab:dataset_config} summarizes the dataset configurations and split sizes for the forecasting tasks in Section~\ref{sec:experimental_setup}.

\begin{table}[H]
\appendixtablestyle
\caption{Dataset configurations. Spatial resolution refers to the source radar grids, and frame size denotes the model input and output size. Forecast frames are shown as observed $\rightarrow$ predicted; split sizes count sequences.}
\label{tab:dataset_config}
\begin{tabular}{@{}>{\raggedright\arraybackslash}p{0.26\linewidth}*{2}{>{\centering\arraybackslash}p{0.30\linewidth}}@{}}
\toprule
\textbf{Setting} & \textbf{SEVIR} & \textbf{MRMS} \\
\midrule
Data variable & VIL & Precipitation rate \\
Spatial / temporal res. & $1~\mathrm{km}$ / $5~\mathrm{min}$ & $0.01^\circ$ / $10~\mathrm{min}$ \\
Forecast frames & $5\rightarrow20$ & $4\rightarrow20$ \\
Frame size & $128\times128$ & $256\times256$ \\
Train / validation / test & 59,530 / 28,145 / 7,220 & 6,807,528 / 12,000 / 12,000 \\
\bottomrule
\end{tabular}
\end{table}

\textbf{SEVIR}~\citep{veillette2020sevir} contains spatiotemporal observations of weather events across the United States, each covering $384~\text{km}\times384~\text{km}$. Following prior studies~\citep{gao2022earthformer,gao2024prediff}, we use the VIL modality, which has a native spatial resolution of 1~km and a temporal resolution of 5~minutes. Following DiffCast and AlphaPre~\citep{yu2024diffcast,lin2025alphapre}, we downsample the frames to $128\times128$ and predict 20 future frames from 5 observed frames. Training includes events up to January 1, 2019. Validation uses later events up to October 1, 2019, and testing uses events after that date. Both cutoffs are inclusive for the earlier split and refer to 00:00 UTC. We exclude events with missing VIL data or duplicate VIL records. From each 49-frame event, we extract 25-frame sequences every five frames. Training augmentation uses rotations by multiples of $90^\circ$ and horizontal or vertical flips, applied consistently across all frames in a sequence. Since SEVIR stores VIL as dimensionless encoded values, we report its color bars and evaluation thresholds without physical units~\citep{veillette2020sevir}.

\textbf{MRMS}~\citep{zhang2016mrms} is a composite radar dataset collected over the contiguous United States, spanning $20^\circ$N--$55^\circ$N in latitude and $130^\circ$W--$60^\circ$W in longitude at a spatial resolution of $0.01^\circ$ per grid. Following NowcastNet~\citep{zhang2023skilful}, we use data from 2016--2020 for model development and 2021 for testing. Within 2016--2020, we reserve the first day of each month for validation and use the remaining days for training. After excluding known anomalous timestamps, we take the last 24 frames of each continuous 40-frame sequence sampled at 10-minute intervals. For training, we sample sequences with replacement and select $256\times256$ crops from $512\times512$ patches, using precomputed sequence and spatial weights. Validation and testing use fixed sequence selections and $256\times256$ center crops.

\textbf{Radar and latent normalization.}
For SEVIR, we normalize VIL values by dividing by 255, following DiffCast~\citep{yu2024diffcast}.
For MRMS, we cap rain rates at $128~\mathrm{mm\,h}^{-1}$ following NowcastNet~\citep{zhang2023skilful}, then apply logarithmic normalization to $[-1,1]$.
For both datasets, we standardize VAE latents using dataset-specific means and standard deviations, building on latent rescaling in latent diffusion models~\citep{rombach2022high}.

\subsection{Evaluation Metrics and Aggregation}
\label{app:evaluation_metrics}
For meteorological skill, we use the Critical Success Index (CSI) and Heidke Skill Score (HSS)~\citep{veillette2020sevir, yu2024diffcast, zhang2023skilful}:
\begin{equation}
\mathrm{CSI}_\tau = \frac{\mathrm{TP}_\tau}{\mathrm{TP}_\tau + \mathrm{FP}_\tau + \mathrm{FN}_\tau},
\end{equation}
and
\begin{equation}
\mathrm{HSS}_\tau = \frac{2(\mathrm{TP}_\tau \cdot \mathrm{TN}_\tau - \mathrm{FP}_\tau \cdot \mathrm{FN}_\tau)}{(\mathrm{TP}_\tau + \mathrm{FN}_\tau)(\mathrm{FN}_\tau + \mathrm{TN}_\tau) + (\mathrm{TP}_\tau + \mathrm{FP}_\tau)(\mathrm{FP}_\tau + \mathrm{TN}_\tau)}.
\end{equation}
Here, $\mathrm{TP}_\tau$, $\mathrm{FP}_\tau$, $\mathrm{FN}_\tau$, and $\mathrm{TN}_\tau$ denote pixelwise true-positive, false-positive, false-negative, and true-negative counts at threshold $\tau$, aggregated over the evaluation set at each lead time. We average CSI and HSS uniformly over lead times and evaluation thresholds.

For perceptual quality, we use LPIPS~\citep{zhang2018unreasonable} and SSIM~\citep{wang2004SSIM}:
\begin{equation}
\mathrm{LPIPS}(\hat{\mathbf{x}},\mathbf{x}^*)
=\sum_{\ell}\frac{1}{H_\ell W_\ell}\sum_{h,w,c}
a_{\ell c}\big(\phi_\ell(\hat{\mathbf{x}})_{hwc}-\phi_\ell(\mathbf{x}^*)_{hwc}\big)^2,
\end{equation}
where $\phi_\ell$ denotes channel-normalized features at layer $\ell$, $a_{\ell c}$ are learned channel weights, and $H_\ell,W_\ell$ are the feature-map dimensions. SSIM is defined as
\begin{equation}
\mathrm{SSIM}(\hat{\mathbf{x}}, \mathbf{x}^*) = \frac{(2\mu_{\hat{x}}\mu_{x^*} + c_1)(2\sigma_{\hat{x}x^*} + c_2)}{(\mu_{\hat{x}}^2 + \mu_{x^*}^2 + c_1)(\sigma_{\hat{x}}^2 + \sigma_{x^*}^2 + c_2)},
\end{equation}
where $\mu$, $\sigma^2$, and $\sigma_{\hat{x}x^*}$ denote local means, variances, and covariance, and $c_1,c_2$ are stabilization constants. Perceptual scores are spatially averaged within each frame and then uniformly averaged over forecast frames and test sequences. Higher CSI, HSS, and SSIM and lower LPIPS indicate better performance.


\section{Additional Experimental Results}
\label{app:additional_results}

\subsection{Forecast Skill Across Lead Times}
\label{app:lead_time_results}
\begin{figure}[htbp]
    \centering
    \includegraphics[width=0.98\textwidth]{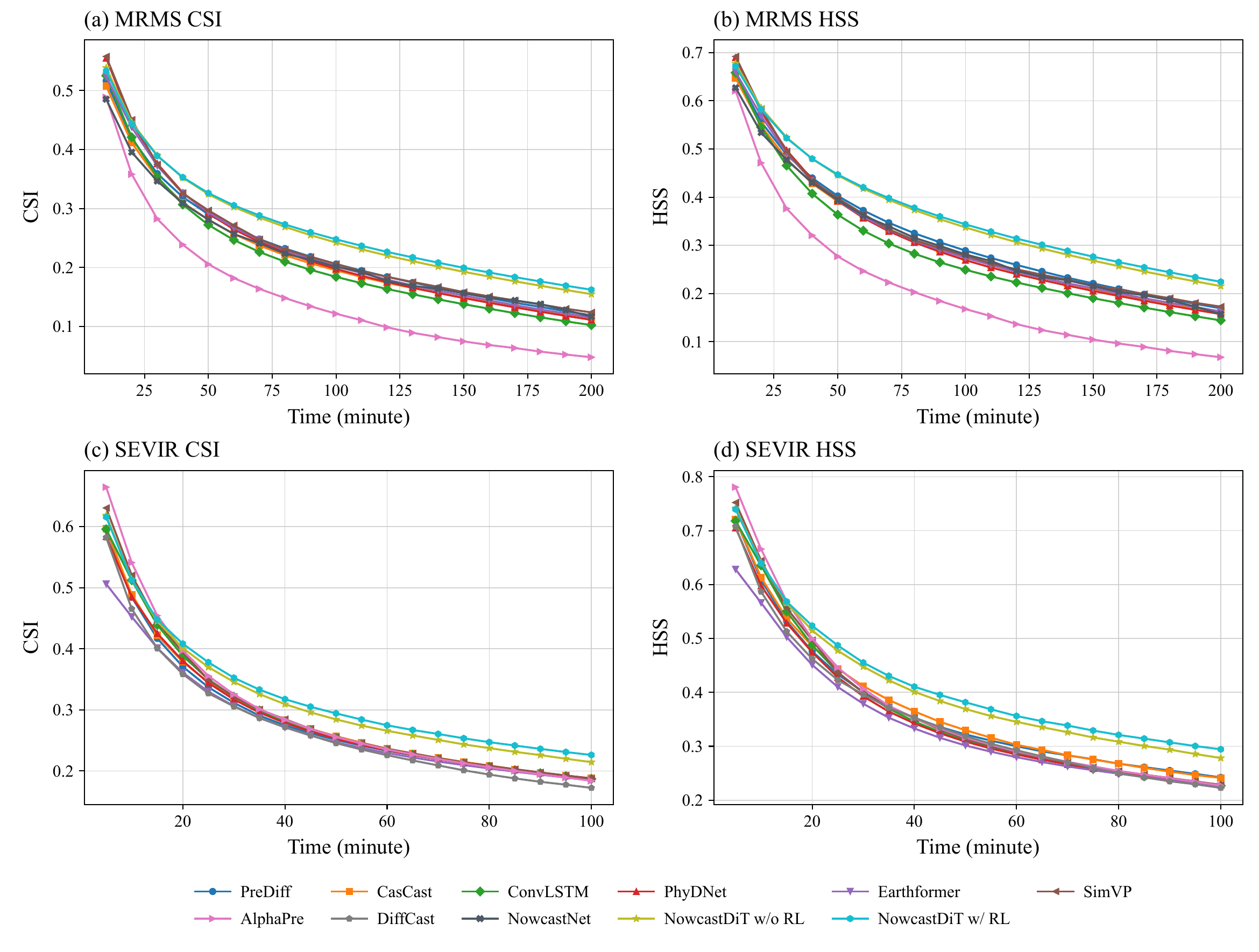}
    \caption{\textbf{Forecast skill across lead times on MRMS and SEVIR.}
    The horizontal axes show forecast lead time in minutes; higher CSI and HSS indicate better performance.}
    \label{fig:nowcastdit_comparison}
\end{figure}

Figure~\ref{fig:nowcastdit_comparison} compares CSI and HSS across forecast lead times on MRMS and SEVIR, complementing the aggregate results in Table~\ref{tab:main}. The curves cover 10--200 minutes on MRMS and 5--100 minutes on SEVIR and include NowcastDiT both with and without RL post-training.

\begingroup
\setlength{\intextsep}{10pt plus 2pt minus 2pt}
\setlength{\textfloatsep}{12pt plus 2pt minus 2pt}
\subsection{Cumulative Component Ablation}
\label{app:full_component_ablation}
Table~\ref{tab:ablation} reports a cumulative ablation on SEVIR, starting from the standard DiT baseline and successively adding 3D RoPE and QK-Norm, DyPro, and timestep-aware RL. Each row retains all components introduced in the preceding rows.
\begin{table}[!htbp]
\appendixtablestyle
\caption{
Cumulative component ablation on SEVIR.
$\uparrow$ indicates higher is better and $\downarrow$ indicates lower is better.
Bold and underlined values denote the best and second-best results, respectively.
}
\label{tab:ablation}
\begin{tabular}{@{}lcccccc@{}}
\toprule
\textbf{Method} & \textbf{CSI}$\uparrow$ & \textbf{CSI-181}$\uparrow$ & \textbf{CSI-219}$\uparrow$ & \textbf{HSS}$\uparrow$ & \textbf{LPIPS}$\downarrow$ & \textbf{SSIM}$\uparrow$ \\
\midrule
DiT & 0.2926 & 0.0948 & 0.0555 & 0.3733 & 0.1545 & 0.7048 \\
+ 3D RoPE,\,QK-Norm & 0.3012 & 0.1082 & 0.0676 & 0.3647 & \textbf{0.1317} & 0.6812 \\
+ DyPro            & \underline{0.3195} & \underline{0.1204} & \textbf{0.0752} & \underline{0.4092} & 0.1486 & \underline{0.7099} \\
+ RL               & \textbf{0.3240}
           & \textbf{0.1241}
           & \underline{0.0750}
           & \textbf{0.4148}
           & \underline{0.1469}
           & \textbf{0.7197} \\
\bottomrule
\end{tabular}
\end{table}

\FloatBarrier
\subsection{CFG and Skill-Aware Post-Training}
\label{app:component_ablation}
We adopt Domain Guidance (DoG)~\citep{zhong2025domain} to integrate CFG with RL post-training, as detailed in Appendix~\ref{app:domain_guidance}.
Table~\ref{tab:rl_cfg_ablation} reports the individual and combined effects of CFG and RL. The combined configuration has the highest aggregate CSI, HSS, and SSIM among these four settings, while CFG alone has the lowest LPIPS, illustrating that the metrics need not improve together.
\begin{table}[!htbp]
\appendixtablestyle
\caption{
Effects of RL and classifier-free guidance on SEVIR.
$\uparrow$ indicates higher is better and $\downarrow$ indicates lower is better.
Bold and underlined values denote the best and second-best results, respectively.
}
\label{tab:rl_cfg_ablation}
\begin{tabular}{@{}cccccccc@{}}
\toprule
\textbf{RL} & \textbf{Guidance} & \textbf{CSI}$\uparrow$ & \textbf{CSI-181}$\uparrow$ & \textbf{CSI-219}$\uparrow$ & \textbf{HSS}$\uparrow$ & \textbf{LPIPS}$\downarrow$ & \textbf{SSIM}$\uparrow$ \\
\midrule
\ding{56} & \ding{56} & 0.2934 & 0.0956 & 0.0560 & 0.3749 & 0.1541 & 0.7041 \\
\ding{56} & \ding{52} & \underline{0.3195} & \underline{0.1204} & \textbf{0.0752} & \underline{0.4092} & \underline{0.1486} & 0.7099 \\
\ding{52} & \ding{56} & 0.3109 & 0.1050 & 0.0588 & 0.3961 & 0.1580 & \underline{0.7156} \\
\ding{52} & \ding{52} & \textbf{0.3240}
           & \textbf{0.1241}
           & \underline{0.0750}
           & \textbf{0.4148}
           & \textbf{0.1469}
           & \textbf{0.7197} \\
\bottomrule
\end{tabular}
\end{table}

\FloatBarrier
\subsection{DyPro Correlation Strength}
\label{app:dypro_results}
Table~\ref{tab:dypro_alpha} reports the SEVIR sensitivity to $\alpha$ across meteorological and perceptual metrics. Among the listed settings, $\alpha=0.5$ gives the highest aggregate CSI. This experiment varies the noise correlation strength. All experiments in this ablation are conducted without RL post-training to isolate the effect of DyPro.
\begin{table}[!htbp]
\appendixtablestyle
\caption{
DyPro correlation strength $\alpha$ on SEVIR.
$\alpha = 0$ corresponds to the standard i.i.d. Gaussian noise prior. 
$\uparrow$ indicates higher is better and $\downarrow$ indicates lower is better.
Bold and underlined values denote the best and second-best results, respectively.
}
\label{tab:dypro_alpha}
\begin{tabular}{@{}ccccccc@{}}
\toprule
$\boldsymbol{\alpha}$ & \textbf{CSI}$\uparrow$ & \textbf{CSI-181}$\uparrow$ & \textbf{CSI-219}$\uparrow$ & \textbf{HSS}$\uparrow$ & \textbf{LPIPS}$\downarrow$ & \textbf{SSIM}$\uparrow$ \\
\midrule
0.00            & 0.3012 & 0.1082 & 0.0676 & 0.3647 & \textbf{0.1317} & 0.6812 \\
0.25            & 0.3188 & 0.1184 & \textbf{0.0759} & 0.4075 & 0.1485 & 0.7090 \\
0.33            & \underline{0.3191} & 0.1192 & 0.0742 & 0.4080 & \underline{0.1483} & 0.7097 \\
0.50            & \textbf{0.3195} & \textbf{0.1204} & \underline{0.0752} & \textbf{0.4092} & 0.1486 & \textbf{0.7099} \\
0.66            & 0.3185 & \underline{0.1194} & 0.0740 & 0.4077 & 0.1495 & \underline{0.7098} \\
0.75            & 0.3184 & 0.1184 & 0.0750 & \underline{0.4084} & 0.1494 & 0.7092 \\
\bottomrule
\end{tabular}
\end{table}

\FloatBarrier
\endgroup

\subsection{Timestep-aware reward variants}
\label{app:reward_ablation}
\begin{figure}[htbp]
    \centering
    \includegraphics[width=\linewidth]{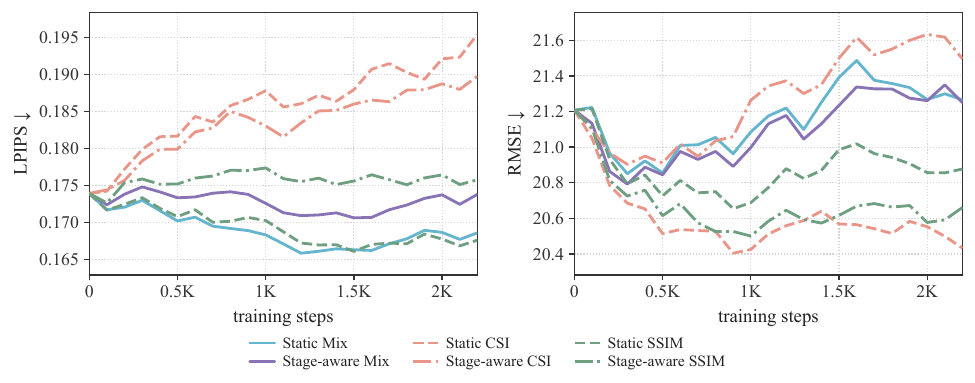}
    \caption{\textbf{Additional reward ablations on SEVIR.} LPIPS (left) and RMSE (right) during RL training for the rewards in Table~\ref{tab:reward_variants}. The legend follows Figure~\ref{fig:analysis_4panel}(d); the timestep-aware perceptual reward uses $0.1(1-t)\mathrm{SSIM}-0.9t\mathrm{LPIPS}$. Lower values indicate better performance for both metrics.}
    \label{fig:reward_ablation_full}
\end{figure}

Table~\ref{tab:reward_variants} defines the six variants in Figures~\ref{fig:analysis_4panel}(d) and~\ref{fig:reward_ablation_full}. Let $C_L$ and $C_H$ denote mean low-threshold pooled CSI and mean high-threshold CSI, using the threshold sets and pooling in Table~\ref{tab:training_configuration}, and let $S=\mathrm{SSIM}$ and $P=\mathrm{LPIPS}$. The mixed components are $M_s=0.5C_L+0.5S$ and $M_d=0.1C_H-0.9P$.
\begin{table}[!htbp]
\appendixtablestyle
\caption{Terminal rewards for the six ablation settings.}
\label{tab:reward_variants}
\begin{tabular}{@{}lcc@{}}
\toprule
\textbf{Reward} & \textbf{Static} & \textbf{Timestep-aware} \\
\midrule
Mixed & $(M_s+M_d)/2$ & $(1-t)M_s+tM_d$ \\
CSI & $(C_L+C_H)/2$ & $(1-t)C_L+tC_H$ \\
Perceptual & $(0.1S-0.9P)/2$ & $0.1(1-t)S-0.9tP$ \\
\bottomrule
\end{tabular}
\end{table}

For the timestep-aware perceptual reward, $R_{\mathrm{noise}}=0.1S$ and $R_{\mathrm{clean}}=-0.9P$, so the reward shifts from SSIM at the noise endpoint to negative LPIPS at the clean endpoint. The static counterpart averages these two endpoint rewards, following the same convention as the CSI and mixed variants. This reward changes the relative weighting of structural agreement and perceptual similarity across denoising transitions. All six variants share the cyclic window schedule in Appendix~\ref{app:mixed_sampling}; the contribution of window scheduling is not separately ablated.

\subsection{Additional Visualizations}
\label{app:additional_visualizations}
\begin{figure}[t]
    \centering
    \includegraphics[width=\linewidth]{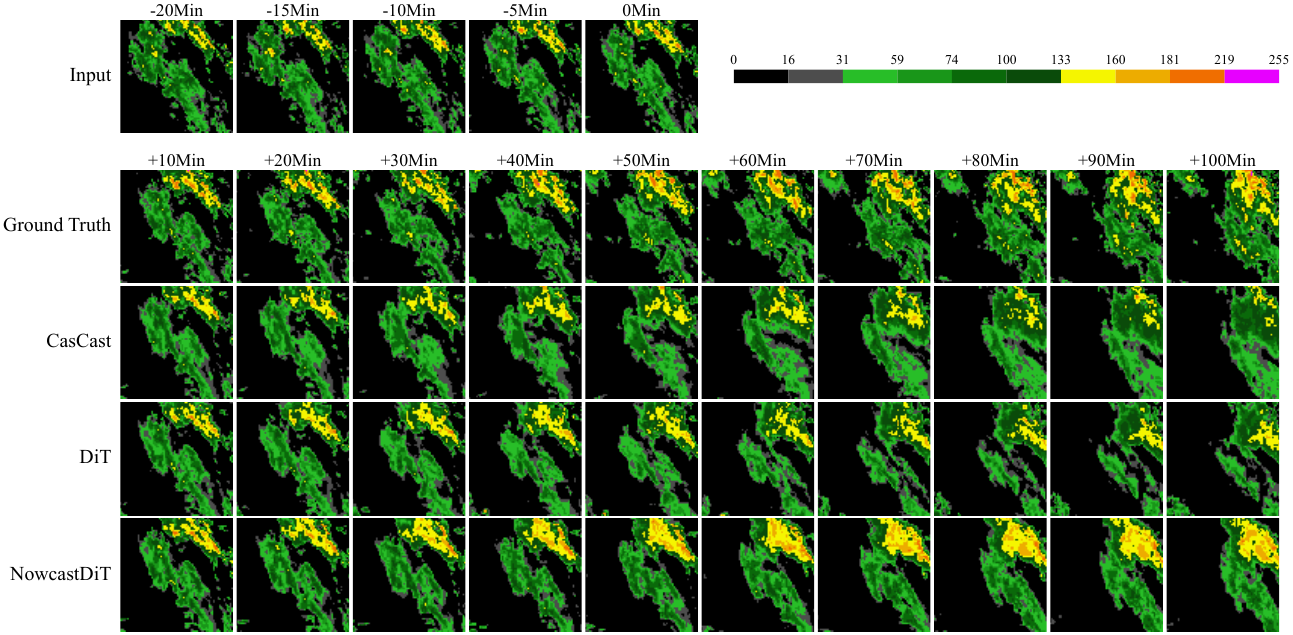}
    \caption{\textbf{Qualitative comparison on SEVIR (case 802).} Five input observations span $-20$ to $0$ minutes at 5-minute intervals. Ground truth and forecasts are shown every 10 minutes up to 100 minutes. Colors indicate VIL values.}
    \label{fig:sevir_case_802}
\end{figure}

\begin{figure}[t]
    \centering
    \includegraphics[width=\linewidth]{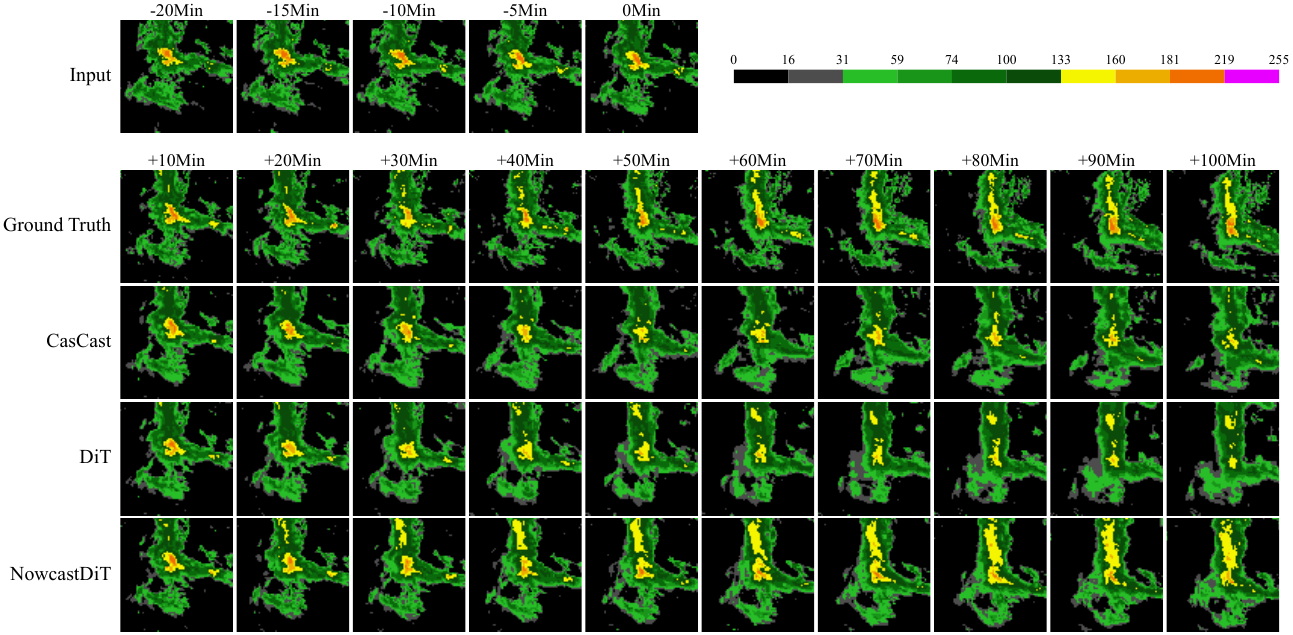}
    \caption{\textbf{Qualitative comparison on SEVIR (case 6155).} Five input observations span $-20$ to $0$ minutes at 5-minute intervals. Ground truth and forecasts are shown every 10 minutes up to 100 minutes. Colors indicate VIL values.}
    \label{fig:sevir_case_6155}
\end{figure}

\begin{figure}[t]
    \centering
    \includegraphics[width=\linewidth]{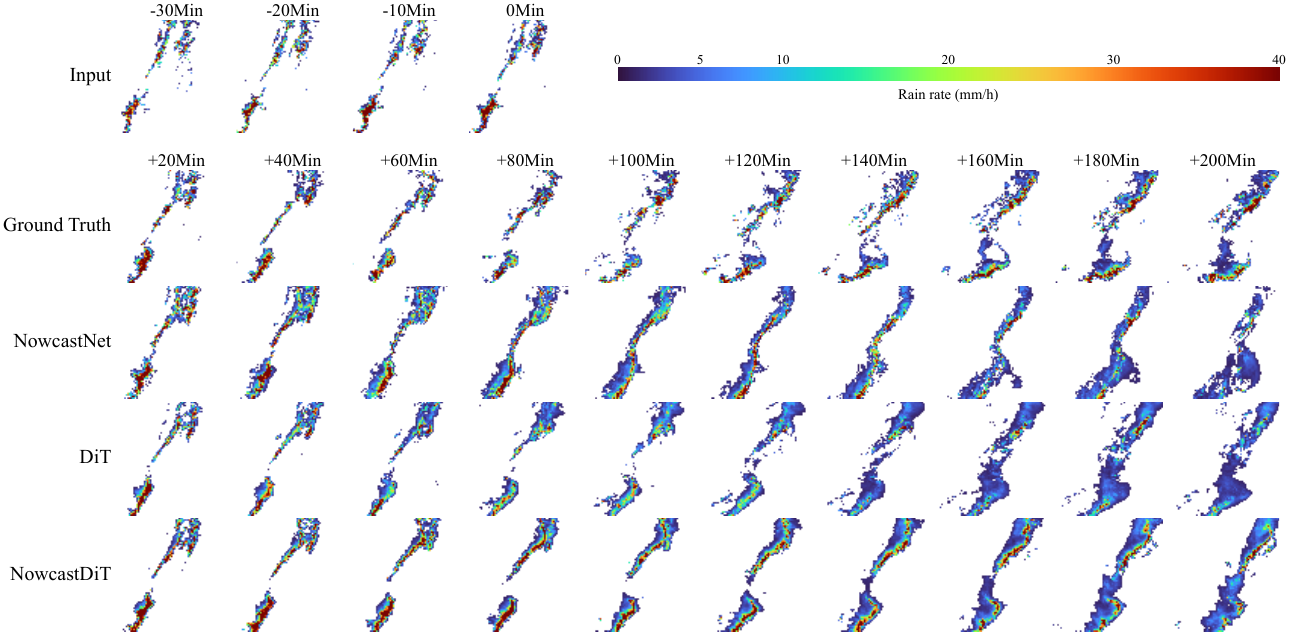}
    \caption{\textbf{Qualitative comparison on MRMS (case 239).} Four input observations span $-30$ to $0$ minutes at 10-minute intervals. Ground truth and forecasts are shown every 20 minutes up to 200 minutes. Colors indicate precipitation rate in mm\,h$^{-1}$.}
    \label{fig:mrms_case_239}
\end{figure}

\begin{figure}[t]
    \centering
    \includegraphics[width=\linewidth]{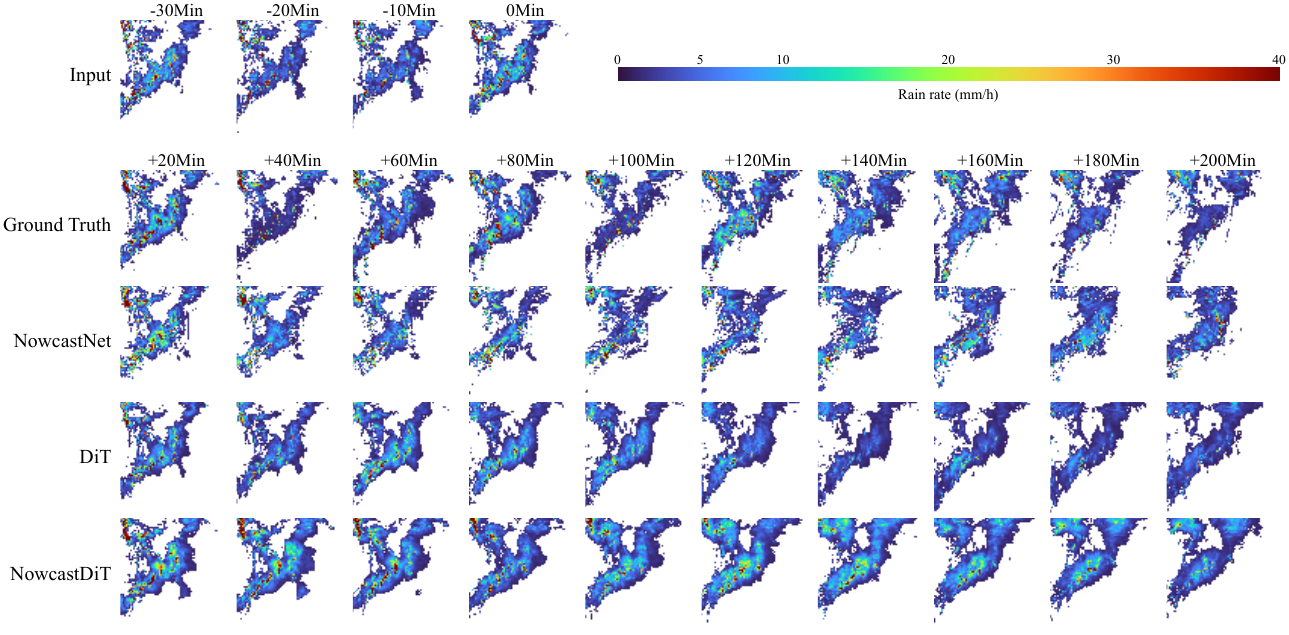}
    \caption{\textbf{Qualitative comparison on MRMS (case 1149).} Four input observations span $-30$ to $0$ minutes at 10-minute intervals. Ground truth and forecasts are shown every 20 minutes up to 200 minutes. Colors indicate precipitation rate in mm\,h$^{-1}$.}
    \label{fig:mrms_case_1149}
\end{figure}

We provide four additional forecast visualizations to complement the qualitative comparisons in the main text. Figures~\ref{fig:sevir_case_802} and~\ref{fig:sevir_case_6155} show SEVIR cases 802 and 6155, comparing NowcastDiT with CasCast and the standard DiT baseline. Figures~\ref{fig:mrms_case_239} and~\ref{fig:mrms_case_1149} show MRMS cases 239 and 1149, comparing NowcastDiT with NowcastNet and the standard DiT baseline. Each figure includes the input observations, ground truth, and forecasts at matched lead times, allowing comparison of precipitation structure and intensity evolution. These selected cases supplement the aggregate evaluation in Table~\ref{tab:main}.

\clearpage

\clearpage

\clearpage

\end{document}